\documentclass{llncs}
\usepackage{makeidx} 

\usepackage{graphicx} 
\usepackage{subfigure}

\usepackage{color}
\usepackage{plain}
\usepackage{rotating}
\usepackage{algorithm}
\usepackage{algorithmic}
\usepackage{amsmath,amsfonts,amssymb,dsfont}

\newcommand{\Drate}{\xi}
\newcommand{\bs}[1]{\boldsymbol{#1}}
\definecolor{orange}{RGB}{255,127,0}
 \definecolor{darkgreen}{rgb}{0.000000,0.392157,0.000000}
  \definecolor{gold}{rgb}{1.000000,0.843137,0.000000}

\newtheorem{assumption}{Assumption}

\newcommand{\x}{\hspace{-1pt}}
\renewcommand{\.}{\hspace{1pt}}

\newcommand{\gleich}{\hspace{5pt}=\hspace{5pt}}

\newcommand{\lit}[1]{\marginpar[\tiny \hfill #1]{\tiny #1}}

\newcommand{\vK}[1]{\lit{(cp. Kojima, Mizuno, Yoshise: A primal-dual interior point algorithm for LP in: Megiddo: Progress in Mathematical Programming\,) }}

\newcommand{\vM}[1]{\lit{(cp.  Mizuno, Todd, Ye: On adaptive step primal-dual interior point algorithms for LP\,) }}
\newcommand{\vYe}[1]{\lit{(cp.  Ye: On the finite convergence of interior point algorithms for LP\,) }}
\newcommand{\vY}[1]{\lit{(cp. Kojima, Mizuno, Yoshise: A  polynomial time algorithm for a class of linear complementary problems\,) }}

\newcommand{\vFo}[1]{\lit{(cp. Forst}}

\newcommand{\R}{\ensuremath{\mathbb{R}}}

\newcommand{\CN}{\ensuremath{{\mathcal{N}}}}

\newcommand{\CW}{\ensuremath{{\mathcal{W}}}}

\newcommand{\tr}{\ensuremath{\operatorname{tr}}}

\newcommand{\ie}{i.\,e.\ }

\newcommand{\PP}%    Primal Problem 
{\[\text{$(P)$}\quad \left\{
\begin{array}{l}
c^Tx\longrightarrow \min\\[0.6ex]
Ax\.=\. b, x  \geq 0\end{array}\right. \]}

\newcommand{\DP}%    Dual Problem 
{\[\text{$(D)$}\quad \left\{
\begin{array}{l}
b^Ty \longrightarrow \max\\[0.6ex]
A^Ty \leq c
\end{array} \right .\]}

\newcommand{\DPs}%    Dual Problem with slack variables 
{\[\text{$(D_e)$}\quad \left\{
\begin{array}{l}
b^Ty \longrightarrow \max\\[0.6ex]
A^Ty + s =  c\\[0.6ex] 
s\geq 0\end{array} \right .\]}

\newcommand{\Pmu}%    Primal logarithmic barrier Problem 
{\[\text{$(P_\mu)$}\quad \left\{
\begin{array}{l}
\Phi_\mu(x) \longrightarrow \min\\[0.6ex]
Ax\.=\. b\end{array}\right. \]}

\newcommand{\Dmu}%    dual logarithmic barrier Problem 
{\[\text{$(D_\mu)$}\quad \left\{
\begin{array}{l}
\widetilde{\Phi}_\mu(y) \longrightarrow \max\\[0.6ex]
A^Ty+s\.=\. c, s  > 0\end{array}\right. \]}

\newcommand{\QP}%    quadratic primal Problem 
{\[\text{$(QP)$}\quad \left\{
\begin{array}{l}
\frac{1}{2}x^TCx+c^Tx\longrightarrow \min\\[0.6ex]
Ax\.\gleich\. b\\x\geq0 \end{array}\right. \]}
\newcommand{\QDe}%    quadratic dual Problem slack
{\[\text{$(QD_e)$}\quad \left\{
\begin{array}{l}
b^Ty-\frac{1}{2}x^TCx\longrightarrow \max\\[0.6ex]
A^Ty-Cx+s\.\gleich\. c\\s\geq0 \end{array}\right.\]}

\newcommand{\QD}%    quadratic dual Problem 
{\[\text{$(QD)$}\quad \left\{
\begin{array}{l}
b^Ty-\frac{1}{2}x^TCx\longrightarrow \max\\[0.6ex]
A^Ty-Cx\.\leq c \end{array}\right. \]}

\newcommand{\Pqmu}%    Primal quadratic logarithmic barrier Problem 
{\[\text{$(QP_\mu)$}\quad \left\{
\begin{array}{l}
\Psi_\mu(x)  \longrightarrow \min\\[0.6ex]
Ax\.=\. b\end{array}\right. \]}

\graphicspath{{./images/}}

\begin{document}

\frontmatter          % for the preliminaries

\pagestyle{headings}  % switches on printing of running heads
\addtocmark{Hamiltonian Mechanics} % additional mark in the TOC

\newcommand{\changedtext}[1]{\textcolor{black}{#1}}
\newcommand{\GR}[1]{\textcolor{red}{\bf #1}}

\mainmatter              % start of the contributions

\title{Probabilistic Clustering of \\
  Time-Evolving Distance Data}

\titlerunning{Probabilistic Clustering of Time-Evolving Distance Data}

\author{Julia E Vogt,\!\inst{1,\ast} Marius Kloft,\!\inst{2} Stefan Stark,\!\inst{1}
Sudhir S Raman,\!\inst{2} \\Sandhya Prabhakaran,\!\inst{4} Volker Roth\inst{4} \and Gunnar R\"atsch\inst{1,\ast} }

\authorrunning{Julia E Vogt et al.} 

\renewcommand*{\thefootnote}{\fnsymbol{footnote}}

\institute{Computational Biology, Memorial Sloan-Kettering Cancer Center, \\1275 York Avenue, New York, NY 10065, USA;
\and
 Department of Computer Science, Humboldt University of Berlin, \\Berlin, Germany;
 \and
Translational Neuromodeling Unit (TNU), Institute for Biomedical Engineering, University of Zurich \& ETH Zurich, Switzerland;
\and
Department of Mathematics and Computer Science, University of Basel, \\Basel, Switzerland
}

\footnotetext{$^\ast$ To whom correspondence should be addressed: \email{vogt@cbio.mskcc.org} and \email{ratschg@mskcc.org}.}

\maketitle              % typeset the title of the contribution

%%%% %%%% %%%% %%%% %%%% %%%% %%%% %%%% Abstract %%%% %%%% %%%% %%%% %%%% %%%% %%%% %%%%

\begin{abstract} 
We present a novel probabilistic clustering model for objects that are
represented via \emph{pairwise distances} \changedtext{and
  observed at different time points.  The proposed method
%groups data at every time point by utilizing 
utilizes the information given by adjacent time points to find the
underlying cluster structure and obtain a smooth cluster evolution.
This approach allows the number of objects and clusters to differ at
every time point, and no identification on the identities of the
objects is needed. Further, the model does not require the number of
clusters being specified in advance -- they are instead determined
automatically using a Dirichlet process prior.  We validate our model
on synthetic data showing that the proposed method is more accurate
than state-of-the-art clustering methods.  Finally, we use our dynamic
clustering model to analyze and illustrate the evolution of brain
cancer patients over time.}
\end{abstract}

%%%% %%%% %%%% %%%% %%%% %%%% %%%% %%%% Introduction%%%% %%%% %%%% %%%% %%%% %%%% %%%% %%%%
\section{Introduction}
A major challenge in data analysis is to find simple representations
of the data that best reveal the underlying structure of the
investigated phenomenon \cite{lee1999learning}. Clustering is a
powerful tool to detect such structure in empirical data, thus making
it accessible to practitioners \cite{jain1988algorithms}.  The problem
of clustering has a very long history in the data mining and machine
learning communities, and numerous clustering algorithms and
applications have been studied in many different scientific
disciplines over the past 50 years \cite{Jain08}.  Applications of
clustering include a large variety of problem domains as, for example,
clustering text, social networks, images, or biomedical data
\cite{bandyopadhyay2003energy,eisen1998cluster,ng2002spectral,steinbach2000comparison}.
Traditional clustering methods such as $k$-means or Gaussian mixture
models \cite{ferguson73}, rely on geometric representation of the
data. Nowadays, however, increasingly often there is no access to an
underlying vectorial representation of the data since only pairwise
similarities or distances are measured.  An example application domain
where such a setting frequently occurs is biomedical data analysis,
where more often than not only \emph{pairwise distance} data is
available, e.g., when DNA or protein sequences are represented as
pairwise distances or string alignments
\cite{string_kernel,LesEskCohWesSta03,RaeSon04,Vert04,SonRaeRie07}.\\[0.5ex]
%Examples of data sets where \emph{pairwise similarity} data is obtained include all types of Mercer kernels \cite{SchSmo02}, be it string alignment kernels over DNA or protein sequences \cite{LesEskNob02,LesEskCohWesSta03,RaeSon04,SonRaeRie07} or diffusion kernels on graphs \cite{Vishwanathan:2010}. 
%Distance data is in no natural way related to the common viewpoint of objects lying in some well behaved space like a Hilbert space. A loss-free embedding of distance  data into a Hilbert or vector space is usually not possible.
%Recently, also probabilistic models for clustering distance data have been proposed, as e.g.~in \cite{Vogt2010} and \cite{fTIWD}. \\[0.5ex]
%Also, learning with distance data is considered a much harder problem than learning with vectorial data, since the inherent structure of $n$ samples is hidden in $n^2$ pairwise relations. \\[0.5ex]
Although many  clustering methods exist that work on distance data, including single linkage clustering, complete linkage clustering, and Ward's clustering \cite{jain1988algorithms}, these methods are \emph{static} methods that are innocuous with respect to a potentially underlying time structure.
%These methods mentioned so far rely on static data and are innocuous with respect to a potentially underlying time structure. 
However, when data is obtained at different points in time, \emph{dynamic} models are needed that take a time component into account. 
For example in cancer research,  genes are frequently measured at different time points,  in order  to examine the  efficiency of a medication over time. 
In Network Security, HTTP connections are recorded at various timestamps, since network behaviors can quickly change over time; in Computer Vision, video streams contain time-indexed sequence of images. % with an additional smoothness over time (timestamps close by are likely to be associated with similar data distributions); in Health Informatics, electronic medical records are associated with a time line that describes disease progression. 
%Also DNA sequences can be analyzed within this framework: in gene finding, commonly subsequences of the genome are processed, the position within the genome can been seen as a ``timestamp''.\\[0.5ex]
To deal with such scenarios, dynamic models that take the evolving nature of data into account are needed.  
Such a requirement has been addressed with evolutionary or dynamic clustering models for \emph{vectorial} data (as for instance in \cite{Ahmed_dynamicnon-parametric}, \cite{Blei_distancedependent}, \cite{TehBluEll2011a}, or \cite{Zoubin05time-sensitivedirichlet}), which  obtain a smooth clustering over multiple time points.
%However, in many practical applications, time series data is only available as scores of pairwise comparisons, especially in biological and medical problems.  
%Surprisingly few machine learning methods exist that work on distance data directly. 
However, to the best of our knowledge, no  time-evolving clustering models exist that work on distance data directly, and clustering of  time-evolving distance data is still an unsolved problem.\\[0.5ex]
\changedtext{In this work we will bridge this gap and present a novel  Bayesian time-evolving clustering model  based on    \emph{distance}  data directly  that is specially tailored to temporal data and does not require direct access to an underlying vector space. Our model will be able to detect cluster popularity over time, based on the rich gets richer phenomenon.   We will be able to make  predictions about how popular a cluster will be at time $t + 1$ if we already knew that it was a rich
cluster at time  point $t$. The assumption that rich clusters get richer seems plausible in many domains,  for instance, a hot
news topic is likely to stay hot for a given time
period.  Our model is also able to cope with variability of data size: the number of data points
may vary between time points, for instance, data
items may arrive or leave. Also, 
the number of
clusters may vary over time
and the model is able to adjust its capacity
accordingly, and automatically. The aim is to find the underlying structure at every time point and  to obtain a smooth cluster evolution which results in an easy  interpretable model. Thereby the information shared across neighboring time points is related to the size of the clusters, the time-varying property of the clusters is assumed to be Markovian, and Markov Chain Monte Carlo (MCMC) sampling is used for inference.\\[0.5ex]
%We present a framework for dynamic clustering of  distance data. 
The presented method is also  applicable for the less general case of  pairwise similarity data, by using a slightly altered likelihood.
Since Mercer kernels can encode similarities between many different kinds of objects (for instance kernels on graphs, images, structures or strings) the method proposed here can in particularly cover the entire scope of applications of kernel-based learning, be it string alignment kernels over DNA or protein sequences \cite{LesEskCohWesSta03,RaeSon04,SonRaeRie07} or diffusion kernels on graphs \cite{Vishwanathan:2010}. %As distance data is the most general data type, we concentrate on distance data in the following.\\[0.5ex]
\\[0.5ex]
We validate our approach by comparing it to baseline methods on simulated  data where  our new model significantly outperforms  state-of-the-art clustering approaches.
%, with accuracy gains up to $15\%-60\%$.
% build the model based on the information of every time point separately, clustering over all time points simultaneously or embedding the distances into a vector space. 
%Important application areas of our approach include all kinds of network prediction problems. 
%We propose and show empirically on simulations that even if an underlying vectorial representation exists, it is of advantage to work directly with the distance matrices to avoid unnecessary bias and variance caused by embeddings (see Section \ref{related_work} for more details).
We apply our novel  model to a highly topical and challenging real world data set of brain cancer patients from Memorial Sloan Kettering Cancer Center (MSKCC). This data consists of clinical notes as part of electronic health records (EHR) of brain cancer patients over 3 consecutive years. We model brain cancer patients over time where patients are grouped together based on the similarity of sentences in the clinical notes (see Section~\ref{sec:EHR}).}
%%By computing the pairwise distances between the inverse covariance matrices that describe the graph structure within one time point, a time series of distance data is obtained. 
%We use our proposed method to  examine which networks behave similar over time under the influence of certain external stimuli,  and obtain interesting results about the temporal response patterns of proteins to a set of external stimuli.\\[0.5ex]
%This paper is structured as follows: in Section~\ref{related_work}, we  recap some essential background knowledge which is necessary  for deriving our time-evolving model, where we closely lean on the notation in \cite{Vogt2010}. In Section~\ref{model} we present our novel time-evolving clustering model. Section~\ref{experiments} contains various synthetic experiments to evaluate the new model as well as a real-world biomedical  experiment on the DREAM breast cancer data set. We conclude with Section~\ref{conclusion}. 

%%%% %%%% %%%% %%%% %%%% %%%% %%%% %%%%Background%%%% %%%% %%%% %%%% %%%% %%%% %%%% %%%%

\section{Background}

\label{related_work}
 In this section we recap important  background knowledge which is essential for the remainder of this paper. 
\vspace{-0.6cm}
\paragraph{Partition Process:}

\begin{figure}
\centering
\includegraphics[width=8cm]{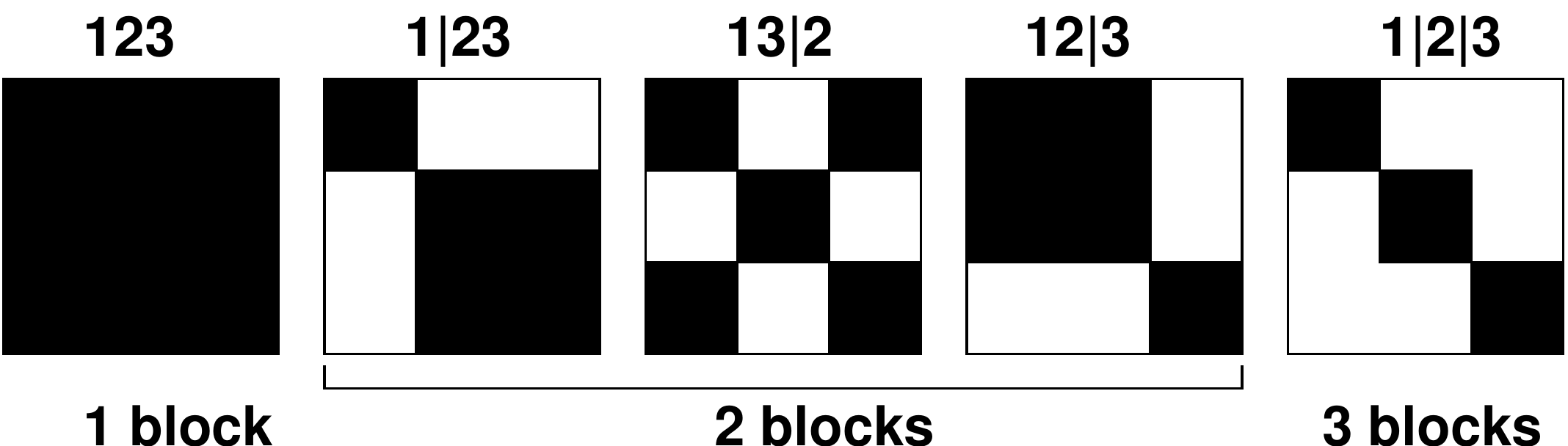}
\caption{\label{fig:lattice}Partition lattice for $\mathbb{B}_3$.}
\end{figure}

Let  $\mathbb{B}_n$ denote a set of partitions of  $[n]$, and $[n] := \{1,\dots,n\}$ denote an index set.
 A partition $B \in \mathbb{B}_n$ is an equivalence relation $B: [n] \times [n] \to \{0,1\}$ with  $B(i,j) = 1$ if $y(i) = y(j)$ and $B(i,j) = 0$ otherwise. $y$ denotes a function that maps $[n]$ to some label set $\mathbb{L}$. Alternatively, $B$ may be represented as a set of disjoint non-empty subsets called ``blocks''.
A \emph{partition process} is a series of distributions $P_n$ on the set $\mathbb{B}_n$ in which $P_n$ is the marginal distribution of $P_{n+1}$. \changedtext{This means, that for each partition $B\in \mathbb{B}_{n+1}$, there exists a corresponding partition $B^*\in \mathbb{B}_n$ which is obtained by deleting the last row and column from the matrix $B$. The properties of partition processes are in detail discussed in \cite{McCullaghHowMany}.} Such a process is called \emph{exchangeable} if each $P_n$ is invariant under permutations of object indices, see \cite{pitman02} for more details. An example for the partition lattice for $\mathbb{B}_3$ is shown in Fig.~\ref{fig:lattice}.

\vspace{-1cm}
\paragraph{Gauss-Dirichlet Cluster Process:} 
The Gauss-Dirichlet cluster process consists of an infinite sequence of points in $\mathbb{R}^d$, together with a random partition of integers into $k$ blocks. A sequence of length $n$ can be sampled as follows \cite{MacEachern94,McCullaghHowMany}: fix the number of mixture modes $k$, generate mixing proportions $\pi = (\pi_1,\dots,\pi_k)$ from a symmetric Dirichlet distribution Dir$(\Drate/k,\dots, \Drate/k)$, generate a label sequence $\{y(1),\dots, y(n)\}$ from a multinomial distribution and forget the labels introducing the random partition $B$ of $[n]$ induced by $y$. Integrating out $\pi$, one arrives at a Dirichlet-Multinomial prior over partitions
\begin{equation}
  \label{eq:dirMultiP}
  P_n(B|\Drate,k) = \frac{k!}{(k-k_B)!} \frac{\Gamma(\Drate) \prod_{b\in B} \Gamma(n_b + \Drate/k)}{\Gamma(n+\Drate)[\Gamma(\Drate/k)]^{k_B}},
\end{equation}
where  $k_B\leq k$ denotes the number of blocks present in the partition $B$ and  $n_b$ is the size of  block $b$.
  The limit as $k\to \infty$ is well defined and known as the Ewens process (a.k.a.~Chinese Restaurant process (CRP)), see for instance \cite{Ewens72,Neal00,blei06}.
Given such a partition $B$, a sequence of $n$-dimensional observations $\bs x_i \in \mathbb{R}^n,\; i = 1,\dots, d$, is arranged as columns  of the $(n \times d)$ matrix  ${X}$, and this $X$ is generated from a zero-mean Gaussian distribution with covariance matrix
\begin{equation}
\label{cov_eq}
\widetilde{\Sigma}_B = I_n \otimes \Sigma_0 +  B \otimes \Sigma_1,
 \quad \text{ with } \quad \text{cov}({X}_{ir},{X}_{js}| B) = \delta_{ij}{\Sigma_0}_{rs} + B_{ij} {\Sigma_1}_{rs} .
\end{equation} 
 $\Sigma_0$  denotes the $(d \times d)$  within-class covariance matrix and $\Sigma_1$ the  $(d \times d)$ between-class matrix, respectively, and $\delta_{ij}$ denotes the Kronecker symbol. 
 Since the partition process is invariant under permutations, we can always think of $B$ being  block-diagonal. 
For spherical covariance matrices (i.e.~scaled identity matrices),  $\Sigma_0 = \alpha I_d, \Sigma_1 = \beta I_d$, the covariance structure reduces to
%\begin{equation}
%\begin{split}
%\label{cov_eq_2}
$\textstyle
\widetilde{\Sigma}_B = I_n \otimes \alpha I_d +  B \otimes \beta I_d\\
= (\alpha I_n + \beta B) \otimes I_d =: \Sigma_B   \otimes I_d,  \text{ with } \text{cov}({X}_{ir},{X}_{js}| B) = (\alpha \delta_{ij} + \beta B_{ij}) \delta_{rs}.$
%\end{split}
%\end{equation}
Thus, the columns of ${X}$ contain independent $n$-dimensional vectors $\bs x_i \in \mathbb{R}^n$ distributed according to a normal distribution with covariance matrix
\begin{equation}
\label{static_cov}
\Sigma_B = \alpha I_n + \beta B.
\end{equation}
Further, the distribution
factorizes over the blocks $b \in B$. Introducing the  symbol ${i_b} := \{i:i\in b\}$ defining an index-vector of all objects assigned to block $b$,  the joint distribution reads 
\begin{equation}
%\begin{split}
\textstyle p(X,B|\alpha,\beta,\Drate,k) =  P_n(B|\Drate,k)
\textstyle \cdot \left[ \prod_{b \in B} \prod_{j=1}^d \mathcal{N}(X_{{i_b} j}| \alpha I_{n_b} + \beta \bs 1_{n_b} \bs 1^t_{n_b}) \right], 
%\end{split} 
\end{equation}
where $n_b$ is the size of block $b$ and $\bs 1_{n_b}$ a $n_b$-vector of ones.
\changedtext{In the viewpoint of clustering, $n$ denote the number of objects we want to partition, and $d$ the dimension of each object}. 
\vspace{-2mm}
\paragraph{Wishart-Dirichlet Cluster Process:} 
Assume that the  random matrix ${X}_{n \times d}$ follows the zero-mean Gaussian distribution specified  in (\ref{cov_eq}), with  $\Sigma_0 = \alpha I_d$ and $\Sigma_1 = \beta I_d$.
Then, conditioned on the partition $B$, the inner product matrix ${K} = {X}{X}^t/d$ follows a (possibly singular) Wishart distribution  in $d$ degrees of freedom, ${K} \sim \mathcal{W}_d(\Sigma_B)$, as was shown in \cite{Srivastava03}.
If we directly observe the dot products $K$, it suffices to consider the conditional probability of partitions $P_n(B|K)$: 
\begin{equation}
\label{eq:factorization}
\textstyle
\begin{split}
&   \textstyle  P_n(B|K,\alpha,\beta,\Drate,k) \propto 
\mathcal{W}_d(K|\Sigma_B)  \cdot P_n(B|\Drate,k)\\
& \textstyle  \propto 	\left|{\Sigma_B}\right|^{-\frac{d}{2}} \exp\left(- \frac{d}{2}{\rm tr}(\Sigma_B^{-1}K)\right) \cdot P_n(B|\Drate,k)   
%  =\frac{|S|^{(d-n-1)/2}}{2^{d n} \Gamma_{n}(d/2)} 
\end{split} 
 \end{equation}

\vspace{-0.5cm}

\paragraph{Information Loss:}

\begin{figure}[h!]
\centering
\includegraphics[width=8.5cm]{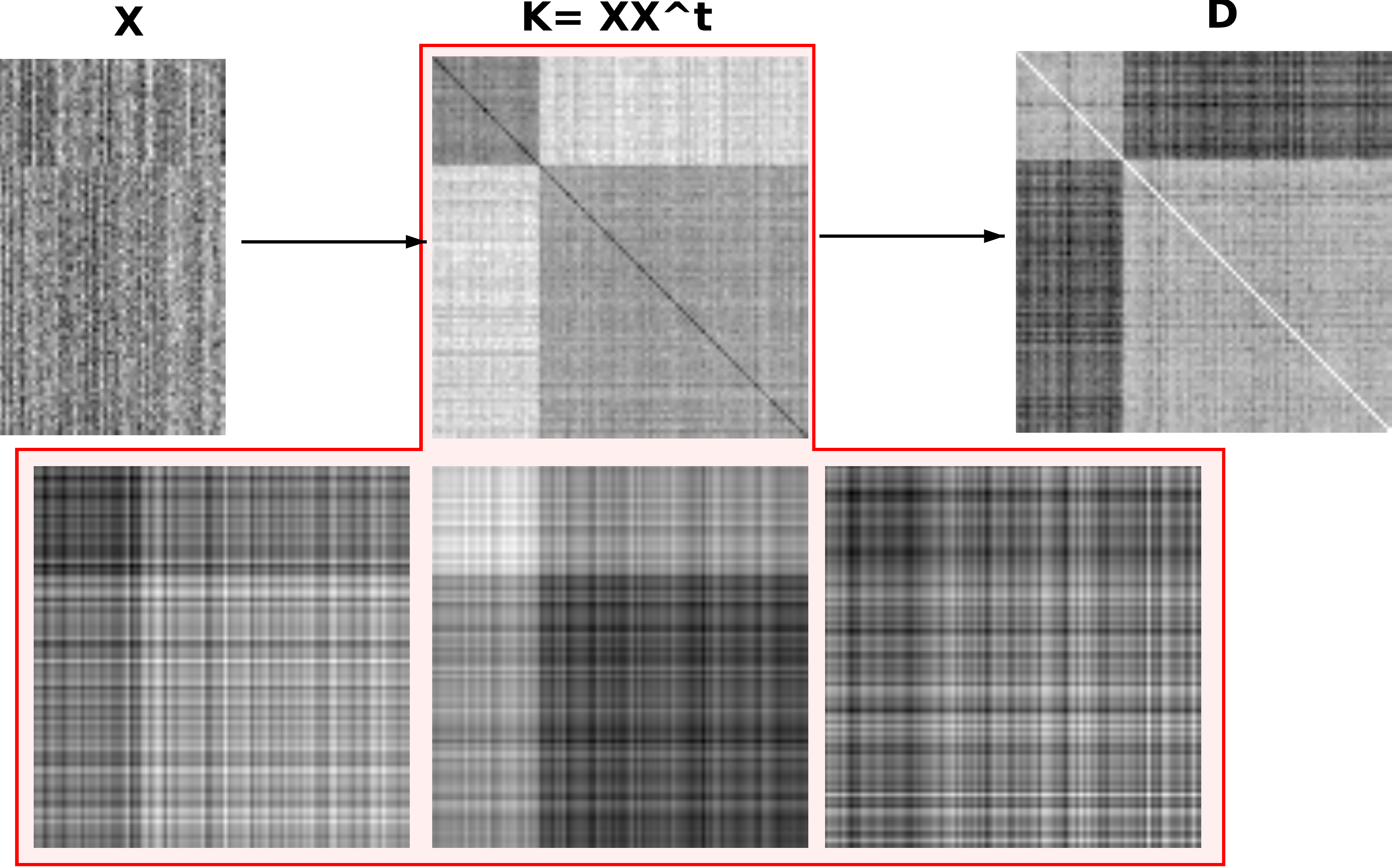}
\caption{\label{fig:information_loss}Information loss that occurs by moving from vectors $X$  to pairwise distances $D$. By moving from $X$ to pairwise similarities $K$, information about rotation of the vectors is lost, by moving from $K$ to $D$, information about translation is lost. One can reconstruct a whole equivalence class of $K$ matrices (four examples are bordered in red) from one distance matrix $D$, \ie the reconstruction of a similarity matrix  $K$ from $D$ is \emph{not} unique, as there is a surjective mapping from a set of $K$ matrices to $D$.}
\end{figure}

\changedtext{Note that we assumed that there exists a matrix $X$ with
  ${K} = {X}{X}^t/d$ such that the \emph{columns} of $X$ are
  independent copies drawn from a zero-mean Gaussian in
  $\mathbb{R}^n$: $\bs x \sim N(\bs \mu = \bs 0_n, \Sigma =
  \Sigma_B)$. This assumption is crucial, since general mean vectors
  correspond to a \emph{non-central} Wishart model
  \cite{Anderson46noncentral}, which can be calculated analytically
  only in special cases, and even these cases have a very complicated
  form which imposes severe problems in deriving efficient inference
  algorithms.\\[0.5ex] By moving from vectors $X$ to pairwise
  similarities $K$ and from similarities to pairwise distances $D$,
  there is a lack of information about geometric transformations:
  assume we only observe $K$ without access to the vectorial
  representations $X_{n\times d}$. Then we have lost the information
  about orthogonal transformations $X\leftarrow X O$ with $OO^t =
  I_d$, i.e.~about rotations and reflections of the rows in $X$. If we
  only observe $D$, we have additionally lost the information about
  translations of the rows $X \leftarrow X + (\bs 1_n \bs v^t + \bs
  v\bs 1_n^t) , \; \bs v \in\mathbb{R}^d$.\\[0.5ex] The models above
  imply that the means in each row are expected to converge to zero as
  the number of replications $d$ goes to infinity. Thus, if we had
  access to $X$ and if we are not sure that the above zero-mean
  assumption holds, it might be a plausible strategy to subtract the
  empirical row means, $X_{n\times d} \leftarrow X_{n\times d}- (1/d)
  X_{n\times d}\bs 1_d \bs 1_d^t$, and then to construct a candidate
  matrix $K$ by computing the pairwise dot products. This procedure
  should be statistically robust if $d\gg n$, since then the empirical
  means are probably close to their expected values. Such a corrected
  matrix $K$ fulfills two important requirements for selecting
  candidate dot product matrices:\\[0.5ex] First, $K$ should be ``typical''
  with respect to the assumed Wishart model with $\bs \mu = \bs 0$,
  thereby avoiding any bias introduced by a particular choice. Second,
  the choice should be robust in a statistical sense: if we are given
  a second observation from the same underlying data source, the two
  selected prototypical matrices $K_1$ and $K_2$ should be
  similar. For small $d$, this correction procedure is dangerous since
  it can introduce a strong bias even if the model is correct: suppose
  we are given two replications from $N(\bs \mu = \bs 0_n, \Sigma =
  \Sigma_B)$, i.e.~$d=2$. After subtracting the row means, all row
  vectors lie on the diagonal line in $\mathbb{R}^2$, and the cluster
  structure is heavily distorted. \\[0.5ex] Consider now the case
  where we observe $K$ without access to $X$.  For ``correcting'' the
  matrix $K$ just as described above we would need a procedure which
  effectively subtracts the empirical row means from the rows of $X$.\\[0.5ex]}
 \changedtext{Unfortunately, there exists no such matrix
   transformation that operates directly on $K$ without explicit
   construction of $X$. It is important to note that the ``usual''
   centering transformation $K \leftarrow QK Q$ with $Q_{ij} =
   \delta_{ij}-\frac{1}{n}$ as used in kernel PCA and related
   algorithms does not work here: in kernel PCA the rows of $X$ are
   assumed to be i.i.d.~replications in $\mathbb{R}^d$. Consequently,
   the centered matrix $K_c$ is built by subtracting the \emph{column}
   means: $X_{n\times d} \leftarrow X_{n\times d} - (1/n) \bs 1_n \bs
   1_n^t X_{n\times d}$ and $ K_c = XX^t = QKQ$.  Here, we need to
   subtract the \emph{row} means, and therefore it is inevitable to
   explicitly construct $X$, which implies that we have to choose a
   certain orthogonal transformation $O$. It might be reasonable to
   consider only rotations and to use the principal components as
   coordinate axes. This is essentially the kernel PCA embedding
   procedure: compute $K_c = QKQ$ and its eigenvalue decomposition
   $K_c = V \Lambda V^t$, and then project on the principal axes: $X =
   V \Lambda^{1/2}$. The problem with this vector-space embedding is
   that it is statistically robust in the above sense only if $d$ is
   small, because otherwise the directions of the principal axes might
   be difficult to estimate, and the estimates for two replicated
   observations might highly fluctuate, leading to different
   column-mean normalizations. Note that this condition for fixing the
   rotation contradicts the above condition $d\gg n$ that justifies
   the subtraction of the means. Further, column mean normalization
   will change the pairwise dissimilarities $D_{ij}$ (even if the
   model is correct!), and this change can be drastic if $d$ is
   small.\\[0.5ex]}
 \changedtext{The cleanest solution might be to consider the distances
   $D$ (which are either obtained directly as input data, or can be
   computed as $D_{ij} = K_{ii} +K_{jj} -2 K_{ij}$) and to avoid an
   explicit choice of $K$ and $X$ altogether. Therefore, one encodes
   the translation invariance directly into the likelihood, which
   means that the latter becomes constant on all matrices $K$ that
   fulfill $D_{ij} = K_{ii} +K_{jj} -2 K_{ij}$.  The information loss
   that occurs by moving from vectors to pairwise similarities and
   from similarities to pairwise distances is depicted in
   Fig.~\ref{fig:information_loss}.}

\vspace{-6mm}
\paragraph{Translation-invariant Wishart-Dirichlet Cluster Process:} 
A method which works directly on distances has been discussed in  (\cite{fTIWD},\cite{Vogt2010}) as an extension of the Wishart-Dirichlet Cluster Process. These methods cluster \emph{static} distance data, and no  access to vectorial data is required.
The  model presented in \cite{Vogt2010} tackles the problem if  we do not directly observe $K$, but only a matrix of pairwise Euclidean distances $D$. 
%A squared Euclidean distance matrix $D$ is characterized by the property of being of negative type, 
%which means that $x^TDx\leq 0$ for any $x$ with $x^T{\bf{1}}=0.$
In the following, the assumption is that the (suitably pre-processed) matrix $D$ contains \emph{squared  Euclidean distances}  with components 
 \begin{equation}
 \label{equ:dist}
 D_{ij}=K_{ii}+K_{jj}-2K_{ij}.
 \end{equation} 
%  In \cite{McCullaghMargDist} equation (3.2), it was shown  that a pairwise distance matrix $D$ which is of negative type %$ D_{ij}=K_{ii}+K_{jj}-2K_{ij}$
%follows a generalized Wishart distribution with $d$ degrees of freedom.
\changedtext{A squared Euclidean distance matrix $D$ is characterized by the property of being of \emph{negative type}, which means that $ \bs x^t D \bs x  = -\frac{1}{2}\bs x^t K \bs x < 0$ for any  $\bs x: \bs x^t \bs 1 =0$.  This condition is equivalent to the absence of negative eigenvalues in  $  K_c = QKQ = -\frac{1}{2} QDQ$. The distribution of $D$ has been formally studied in \cite{McCullaghMargDist}, Eq.~(3.2), where it was shown that if $K$ follows a standard Wishart generated from an underlying zero-mean Gaussian process, ${K} \sim \mathcal{W}_d(\Sigma_B)$, $-D$ follows a generalized Wishart distribution, $-{D} \sim  \mathcal{W}(\bs 1, 2\Sigma_B) =  \mathcal{W}(\bs 1, -\Delta)$ defined with respect to the transformation kernel $\mathbb{K} = \bs 1$, where   $\Delta_{ij} = {\Sigma_B}_{ii} + {\Sigma_B}_{jj} - 2{\Sigma_B}_{ij}$. To understand the role of the transformation kernel it is useful to  
 introduce the notion of a 
generalized Gaussian distribution with kernel $\mathbb{K} =  \bs 1$: 
$X \sim N( \bs 1, \bs \mu , \Sigma)$. For any transformation $L$ with
$L \bs 1 =0$, the meaning of the general Gaussian notation is: $ LX  \sim N(L \bs\mu, L \Sigma L^t)$.
It follows that under the kernel $\mathbb{K} = \bs 1$, two parameter settings $(\bs \mu_1, \Sigma_1)$ and  $(\bs \mu_2, \Sigma_2)$ are equivalent if $L(\bs \mu_1- \bs \mu_2) = \bs 0 $ and $L(\Sigma_1-\Sigma_2) L^t = 0$, i.e.~if $\bs \mu_1- \bs \mu_2 \in \bs 1$, and 
$(\Sigma_1-\Sigma_2) \in \{\bs 1_n \bs v^t + \bs v \bs 1_n^t: \bs v \in \mathbb{R}^{n} \}$, a space which is usually denoted by  $\text{ sym}^2(\bs 1 \otimes  \mathbb{R}^n)$. It is also useful to introduce the distributional symbol   $K \sim \mathcal{W}(\mathbb{K},\Sigma)$ for the generalized  Wishart distribution of the random matrix  ${K} = XX^t$ when $X \sim N(\mathbb{K}, \bs 0 , \Sigma)$.
The key observation in \cite{McCullaghMargDist} is that  $D_{ij} = K_{ii} +K_{jj} -2 K_{ij}$ defines a linear transformation on symmetric matrices with kernel  $ \text{ sym}^2( \bs 1 \otimes  \mathbb{R}^n)$ which implies that  the distances follow a generalized Wishart distribution with kernel $\bs 1$:   
$-{D} \sim  \mathcal{W}(\bs 1,2\Sigma_B) = \mathcal{W}(\bs 1,-\Delta)$ and 
 \begin{equation}
 \label{equ:delta}
 \textstyle \Delta_{ij}=\Sigma_{B_{ii}}+\Sigma_{B_{jj}}-2\Sigma_{B_{ij}}.
 \end{equation}
  In the multi-dimensional case with spherical within- and between covariances  we generalize the above model to Gaussian random matrices $X \sim N(\bs \mu, \Sigma_B \otimes  I_d)$. Note that the $d$ columns of this matrix are i.i.d.~copies. The distribution of the matrix of squared Euclidean distances $D$ then follows a generalized Wishart with $d$ degrees of freedom $ -{D} \sim  \mathcal{W}_d(\bs 1,-\Delta)$. This distribution differs from a standard Wishart in that the inverse matrix $W = \Sigma_B^{-1}$ is substituted by the matrix $\widetilde{W} = W - (\bs 1^t W \bs 1 )^{-1} W \bs 1 \bs 1^t W$ and the determinant $|\cdot|$ is substituted by a generalized $\det(\cdot)$-symbol which denotes the product of the nonzero eigenvalues of its matrix-valued argument (note that $\widetilde{W} $ is rank-deficient). The conditional probability of a partition then reads
\begin{equation}
\begin{split}
  \label{eq:distrD}
 & P(B|D,\bullet)  \textstyle   \propto  \mathcal{W}(-D|\bs 1,-\Delta) \cdot P_n(B|\Drate,k)\\
&  \textstyle  \propto  \det(\widetilde{W})^\frac{d}{2}  \exp\left(\frac{d}{4} \text{tr}(\widetilde{W} D) \right) \cdot P_n(B|\Drate,k). 
\end{split}
\end{equation}
and  the probability density function (which serves as likelihood function in the model) is then defined as
    \begin{equation}
    \label{equ:D}
 \textstyle  f(D) \propto \det(\widetilde{W})^{\frac{d}{2}}\exp{\left(\frac{d}{4}\tr{(\widetilde{W}D)}\right)}.
    \end{equation}
Note that in spite of the fact that this probability is written as a function of $W = \Sigma_B^{-1}$, it is constant over all choices of $\Sigma_B$ which lead to the same $\Delta$, i.e.~independent under translations of the row vectors in $X$. For the purpose of inferring the partition $B$, this invariance property means that one can simply use a block-partition covariance model $\Sigma_B$ and assume that the (unobserved) matrix $K$ follows a standard Wishart distribution parametrized by $\Sigma_B$. We do not need to care about the exact form of $K$, since the conditional posterior for $B$ depends only on $D$.  Extensive analysis about the  influence of encoding the translation invariance into the likelihood versus the standard WD process and row-mean subtraction was conducted in \cite{Vogt2010}.
 }

%%%% %%%% %%%% %%%% %%%% %%%% %%%% %%%%Te-TIWD%%%% %%%% %%%% %%%% %%%% %%%% %%%% %%%%

\section{A Time-evolving Translation-invariant Wishart-Dirichlet Process}
\label{model}
In this section, we present a  novel dynamic clustering approach, the time-evolving translation-invariant Wishart-Dirichlet process (Te-TiWD) for clustering  distance  data that is available at multiple time points. 
In this model, we assume that  pairwise  distance data $D_t$ with $ 1\leq t \leq T$ is available over $T$ time points. At every time point $t$ all objects are fully exchangeable, and the number of data points may differ at the different time points. This model clusters data points over multiple time points, allowing group memberships and the number of clusters to evolve over time by addition, deletion or change in existing clusters. The model is based on the static clustering model that was proposed in \cite{Vogt2010} which is not able to account for a time structure. 
%A simplified example of distance data obtained at three time points is  given in Fig.~\ref{fig:time series}.   Every time point contains pairwise distances between 20 objects which are grouped into three blocks in the first time point, four blocks in the second and four in the third time point. 
\changedtext{Note that  our model completely ignores any information about the identities of the data points across the time points, which makes it possible to cluster different objects over time.  Table \ref{tab:notations} summarizes notations which we will use in the following sections.}

\renewcommand{\arraystretch}{1.2}
\begin{table}[t]
\small{
\begin{center}
  \begin{tabular}{ |c  || l |}
    \hline
       $D_t$ & distance matrix at time point $t$   (cf.~(\ref{equ:dist})) \\ \hline
          $\Delta_t$ & $\Delta$  matrix at time point $t$ (cf.~(\ref{equ:delta}))\\ \hline
    $B_t$  & partition matrix at time point $t$  \\ \hline
    $k_{b_t}$ & number of blocks $b_t$ present \\
    &in the partition $B_t$ \\ \hline
    $n_{b_t}$ & the size of block $b_t$\\ \hline
    $n_{b_t}^{(-l)}$ & size of block $b_t$ without object $l$\\ \hline
     $n_t$ &number of data points  \\
   &  present at the $t$-th time point\\ \hline
     $A_t$ &$k_{b_t}\times k_{b_t}$ matrix \\ \hline
     $A_{t_{ij}}$ & the  between-class variance of\\
     & block $i$ and block $j$\\ \hline
  $[B_t]_{t=1}^T$& is defined as $(B_1, B_2,...,B_T)$ \\ \hline
  $[A_t]_{t=1}^T$ & is defined as $(A_1, A_2,...,A_T)$\\ \hline
  $p([B_t]_{t=1}^T)$&$=p(B_1)p(B_2|B_1)....p(B_T|B_{T-1})$ \\& defines a first-order Markov chain\\ \hline
  $p([A_t]_{t=1}^T)$&$ =p(A_1)p(A_2|A_1)....p(A_T|A_{T-1})$\\ &defines a first-order Markov chain\\ \hline
%  $p(B_1,B_2,...,B_T)=$& first-order Markov chain\\ \hline
%  $p(A_1,A_2,...,A_t) $=& first-order Markov chain\\ \hline
  $[B]_{t-}$& $B$ matrices at all time points \\
  & except at time point $t$\\ \hline
  \end{tabular}
  \vspace{0.3cm}
  \caption{\label{tab:notations}\changedtext{Notations used throughout this manuscript.}}
\end{center}
}
\end{table}

%%%% %%%% %%%% %%%% %%%% %%%% %%%% %%%%Model%%%% %%%% %%%% %%%% %%%% %%%% %%%% %%%%

 \subsection{The Model}

The aim of the proposed method is to cluster distance data  $D_t$ at multiple time points, for $ 1\leq t \leq T$. For every time point under consideration, $t$, we obtain a distance matrix $D_t$ and we want 
to infer the partition matrix $B_t$,  by utilizing the partitions from adjacent time points.  By using information from adjacent time points, we expect better clustering results than clustering  every time point independently. At every time point, the number of data points may differ, and some clusters may die out or evolve over time.  
The assumptions on the data are the following: 
\begin{assumption}
Given a partition $B_t$, a sequence of the assumed underlying $n_t$-dimensional vectorial  observations $x_{t_i} \in \mathbb{R}^{n_t}$, $i=1,...,d_t$, are arranged as columns of the $(n_t\times d_t)$ matrix $X_t$, i.e.\ $x_{t_1},...,x_{t_{d_t}}  \hspace{1mm}\substack{i.i.d \\ \sim} \hspace{1mm}\CN(0,\Sigma_{B_t})$,
with covariance matrix 
\begin{equation}
\label{dynamic_cov}
\Sigma_{B_t}=\alpha I_{n_t}+\Sigma_{A_t}.
\end{equation}
\end{assumption}

%%%% %%%% %%%% %%%% %%%% %%%% %%%% %%%%Covariance Matrix%%%% %%%% %%%% %%%% %%%% %%%% %%%% %%%%

\begin{figure}[]
  \begin{minipage}{0.24\linewidth}
    \centering
    \includegraphics[height=2cm]{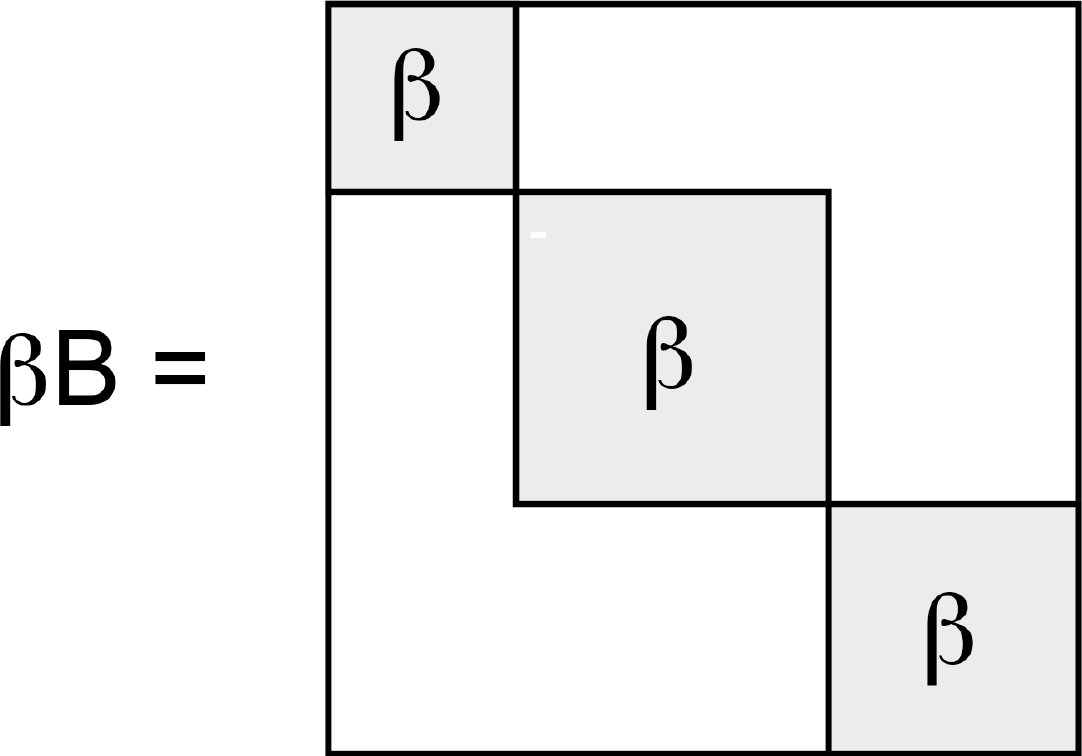}%\vspace{-1cm}
  \end{minipage}
  %\hspace{0.65cm}
  \begin{minipage}{0.24\linewidth}
    \centering
    \includegraphics[height=2.6cm]{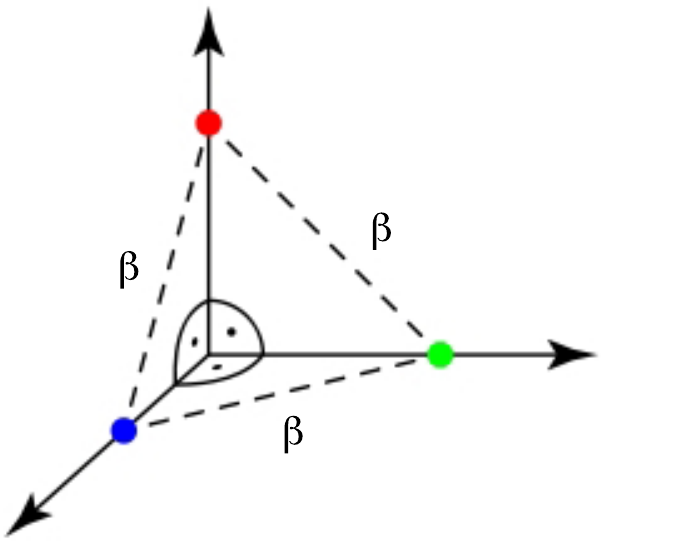}
  \end{minipage}
  \hspace{0.15cm}
  \begin{minipage}{0.24\linewidth}
    \centering
    \includegraphics[height=2cm]{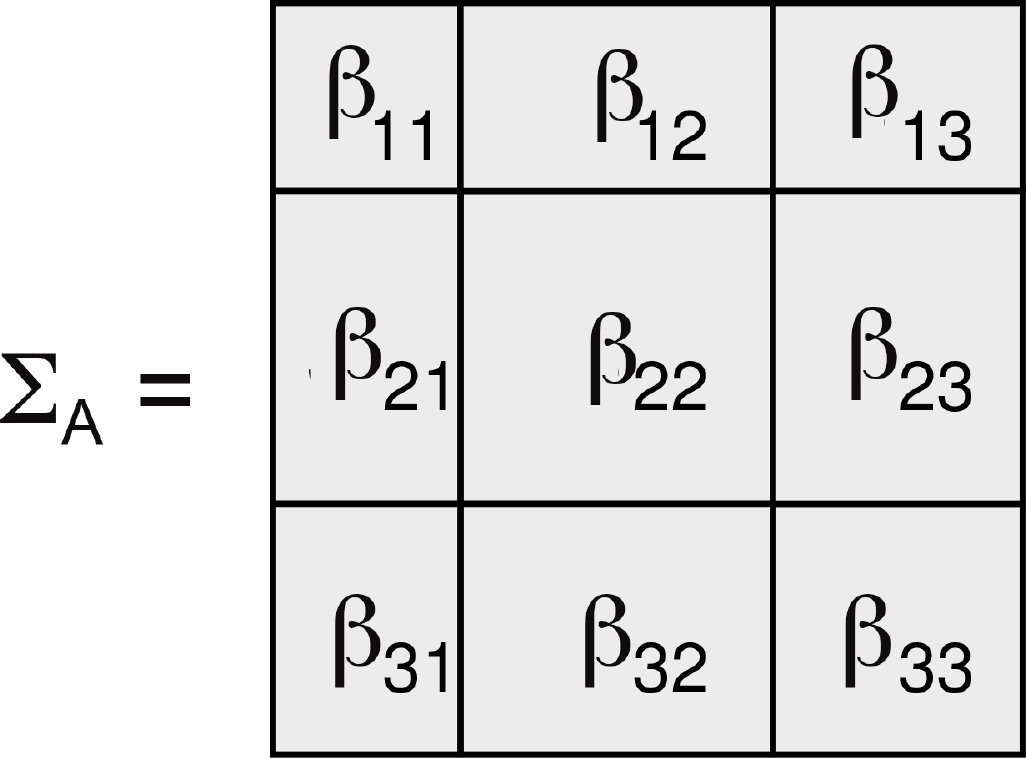}%\vspace{-1cm}
  \end{minipage}
  %\hspace{0.65cm}
  \begin{minipage}{0.24\linewidth}
    \centering
    \includegraphics[height=2.6cm]{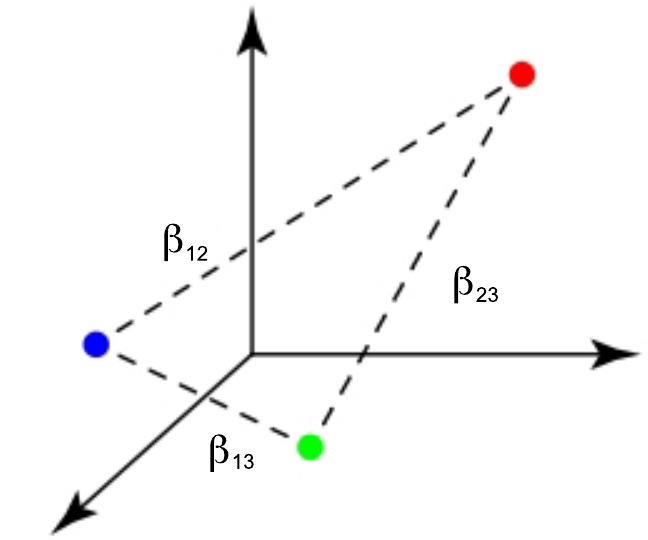}
  \end{minipage}
  % \vspace{-1cm}
  \caption{\label{fig:Sigma_A} Different models for clustering. Left: example of the block diagonal structure of $\beta B$ for three blocks, all cluster centroids must be equidistant. Right: example of the full covariance matrix $\Sigma_{A_t}$ (for better readability, we drop the time index $t$ in the figure), which allows differing distances between cluster centroids.} 
\end{figure}

\paragraph{Covariance matrix $\Sigma_{B_t}$.}
In the static clustering method, the underlying vectorial data was assumed  to be distributed according to a Gaussian distribution with mean $0$, $x_1,...,x_d  \hspace{1mm}\substack{i.i.d \\ \sim} \hspace{1mm}\CN(0,\Sigma_B)$ with $\Sigma_B:=\alpha I_n +\beta B$, (cf.~\eqref{static_cov}), where $\beta B$ describes the \emph{between class} covariance matrix. As $\beta$ denotes a scalar,  all clusters in the static clustering are equidistant (as demonstrated in left of Fig.~\ref{fig:Sigma_A}).  To model  time evolving data, we need a more flexible  between-class covariance matrix $\Sigma_{A_t}$ which allows that cluster centroids  have different distances to each other. These full $\Sigma_{A_t}$ matrices are necessary for a time-evolving clustering model, as the clusters are coupled over the different time-points due to the geometric information of the clusters, and this coupling can only be captured by modeling a richer covariance. 
 Hereby $\Sigma_{A_t} \in \R^{(n_t \times n_t)}$  is obtained 
in the following way 
\begin{equation}
\Sigma_{A_t}=Z_tA_tZ_t^T
\end{equation}
with $Z_t\in\{0,1\}^{n_t\times k_{b_t}}$. The matrix $Z_t$ associates an object with one out of $k_{b_t}$ clusters. As every object can only belong to exactly one cluster,  $Z_t$ has a single element of $1$ per row.
In Fig.~\ref{fig:Sigma_A} we demonstrate examples of $\beta B$ and $\Sigma_{A_t}$ as well as the corresponding cluster arrangements which the matrices imply.  \\
\newpage
Note that  $\Sigma_{A_t}$ is a more general version of $\beta B$:
\begin{lemma}

 $\Sigma_{A_t}=\beta B$ \hspace{0.5cm}iff \hspace{0.5cm} $A_{t_{ij}}=\left\{\begin{array}{cc}
0&\, \, \textnormal {if}  \,  \,  i \neq j\\
\beta \, \, & \textnormal {if} \, \,  i=j
\end{array}
\right . .
$
\end{lemma}

%%%% %%%% %%%% %%%% %%%% %%%% %%%% %%%%prior over B_t%%% %%%% %%%% %%%% %%%% %%%% %%%% %%%%

\paragraph{Prior over the block matrices $B_t$.}
The prior over the block matrices $B_t$ is defined in the following way. The prior for $B_t$ in one epoch is the Dirichlet-Multinomial prior over partitions as in (\ref{eq:dirMultiP}). 
 Using the definition of the conditional prior over clusters as defined in
\cite{Ahmed_dynamicnon-parametric}, we extend this idea to the prior over partitions.
In a generative sense, the same idea is used to generate a labeled set of partitions and then we forget the labels to
get a distribution over partitions. 
 By $n^{t}_{b_{t-1}}$ we denote the size of block ${b_{t-1}}$ if the corresponding block is present at time point $t$ as well.
 \changedtext{We consider the following generative process for a finite dynamic mixture model with $k$ mixtures (cf.~\cite{Ahmed_dynamicnon-parametric}, Eqs.~(4.5), (4.6) and (5.9)): for each time point $t$,  we generate mixing proportions $\pi_t=(\pi_{t1},...,\pi_{tk})$ from a symmetric Dirichlet distribution $\textnormal{Dir}(\xi/k+n_{t-1},...,\xi/k+n_{t-1})$. As in the static case, we generate a label sequence from a multinomial distribution and forget the labels introducing the random partition $B_t$. Integrating out $\pi_t$,  the conditional
distribution for Dirichlet-Multinomial prior over partitions,  given the partitions in the previous time point $(t-1)$, can be written as:
\begin{equation}
\label{t_prior}
 P_{n_t}(B_t | B_{t-1},\xi,k)=  
 \frac{k!}{(k-k_{B_t})!} \frac{\Gamma(\xi + n_{t-1}) \prod_{b_t \in B_t} \Gamma(n^{t}_{b_{t-1}} +
\xi/k + n_{b_t})  }{\Gamma(n_t + \xi + n_{t-1}) 
\prod_{b_t \in B_t} \Gamma(\xi/k + n^{t}_{b_{t-1}})}
\end{equation}
Note that  (\ref{t_prior}) defines a partition process as described in section \ref{related_work} with $P_{n_t}$ being the marginal distribution of $P_{n_{t-1}}$, and it also is an exchangeable process, as each $P_{n_t}$ is invariant under permutation of  object indices.}

%%%% %%%% %%%% %%%% %%%% %%%% %%%% %%%%prior over A_t%%% %%%% %%%% %%%% %%%% %%%% %%%% %%%%

\paragraph{Prior over $A_t$.}
The prior over the $A_t$ matrices is given by a Wishart distribution, 
  $P(A_t|A_{t-1}) \sim \mathcal{W}_d(A_t|A_{t-1})$  and $S_0:=P(A_{1})= \mathcal{W}_d(A_{1}|I_{k_{b_1}})$. The degrees of freedom $d$ influences the behavior of the Wishart distribution: a low value for $d$ allows drastic changes in the clustering structure, a high value for $d$ allows fewer changes. We also have to consider that the size of $A_{t-1}$, $A_{t}$ and $A_{t+1}$ might differ, as it is possible that the number of clusters in every epoch is different. Therefore, we
   consider the following two cases:  
%  \vspace{-2mm}
  \begin{itemize}
  \item[ 1)] \emph{if there are more blocks at time  $t-1$ than at time $t$}, \ie $k_{b_{t-1}}>k_{b_t}$: \\delete corresponding rows and columns in $A_{t-1}$. With $A'_{t-1}$ we denote the ``reduced" matrix. Then it holds that
$A_{t} \sim \mathcal{W}_d(A'_{t-1})$\\
%\vspace{-0.5mm}
\item[ 2)] \emph{if there are fewer blocks at time  $t-1$ than at time $t$}, \ie  if $k_{b_{t-1}}<k_{b_t}$: \\first, draw a $k_{b_{t-1}} \times k_{b_{t-1}}$ matrix $A'_{t}$ from
 $A'_{t} \sim \mathcal{W}_d(A_{t-1})$.
Second, augment as many new rows and columns  as needed to obtain the full positive definite $(k_{b_{t}}) \times (k_{b_{t}})$ matrix $A_{t}$. 
 We can draw the additional rows and columns of $A_t$ in the following way  (see \cite{bilodeau} for details):   
\begin{equation}
A_t=\left(
\begin{matrix}
A_{11} A_{12}\\
A_{21} A_{22}
\end{matrix}
\right)
\end{equation}
%\vspace{-0.5mm}
with $A_{11}:=A'_t  \in \mathbb{R}^{(k_{b_{t-1}}) \times (k_{b_{t-1}})}$, $A_{21}\in \mathbb{R}^{1\times (k_{b_{t-1}})}$,
$A_{12}\in \mathbb{R}^{ (k_{b_{t-1}}) \times 1}$ and $A_{22}\in \mathbb{R}$.
One obtains $A_{12}$,   $A_{21}$ and   $A_{22}$ in the following way:
\begin{eqnarray}
\label{newrow}
\begin{split}
A_{12}|A_{11}&\sim& \mathcal{N}(0,A_{11}\otimes s)\\
A_{22.1}&\sim& \mathcal{W}_1(d-k_{b_t},s)\\%=Gam((df-(k_b+1))/2,1/(2s))\\
A_{22}&=&A_{22.1}+A_{21}A_{11}^{-1}A_{12}
\end{split}
\end{eqnarray} 

where $s$ denotes a scalar value and $d$ the degrees of freedom of the Wishart distribution $\mathcal{W}_d(A_{t-1})$.
\end{itemize}

\begin{figure}[h!]
  \centering \includegraphics[width=9.5cm]{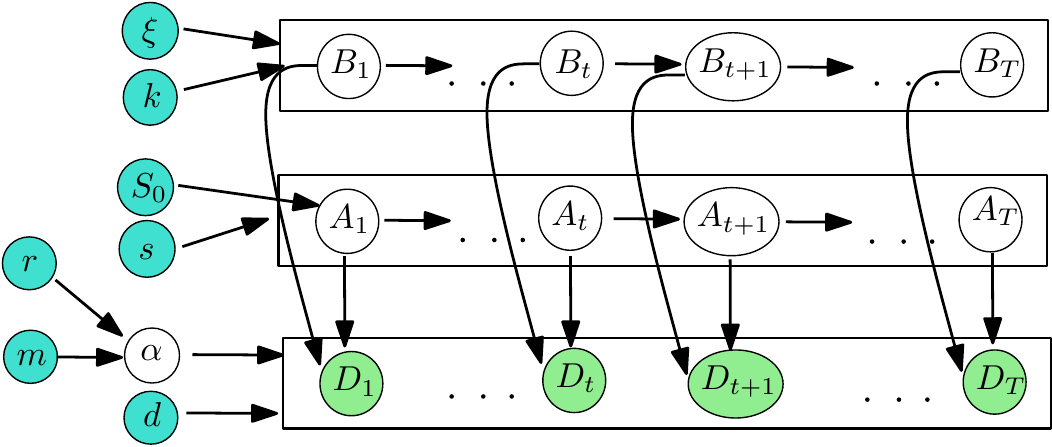}\\
  \caption{\label{fig:gen_model} Depiction of the generative model of Te-TiWD with all hyper-parameters  and parameters. Shaded circles correspond to fixed or observed variables, unshaded to latent variables. Arrows that point to a box mean that the parameters apply to all the variables inside the box, whereas arrows that directly point to a variable only apply to that single variable. \changedtext{$D_t$ denote the distance matrices observed at different points in time, $B_t$ denote the inferred partitions and $A_t$ the between class covariance matrices  at different time points $1\leq t\leq T$. }}
\end{figure}

A graphical depiction of the generative model of Te-TiWD is given in
Fig.~\ref{fig:gen_model}.

\paragraph{Posterior over $B_t$ and $A_t$.}
\changedtext{With the likelihood for every time point, analogous to  Eq.~(\ref{equ:D}), and the prior over  $A_t$ and $B_t$, we can now write down the equations for the posterior over $B_t$ and $A_t$
for all time points $t\in \{1,2,...,T\}$:}
\vspace{-2mm}
\begin{eqnarray}
\label{posterior}
%\begin{split}
 p([B_t]_1^T,[A_t]_1^T | [D_t]_1^T, \bullet) \propto
% p([B_t]_1^T,[A_t]_1^T | [D_t]_1^T, \alpha, \xi, k ) \propto
  \prod_{t=1}^{T}
\CW_d^-(D_t |\textbf{1},\Delta_t) P([B_t]_1^T) P([A_t]_1^T)\\
=\prod_{t=1}^{T} \det(\widetilde{W_t})^{\frac{d_t}{2}}
\exp{\left(\frac{d_t}{4}\tr{(\widetilde{W_t}D_t)}\right)} P([B_t]_1^T) P([A_t]_1^T)
%\end{split}
\end{eqnarray}
with 
  $\widetilde{W_t}:=W_t-(\bs{1}^{T}W_t\bs{1})^{-1}W_t\bs{1}\bs{1}^{T}W_t$,  where  $W_t:=\Sigma_{B_t}^{-1}$ (cf.~\eqref{eq:distrD} and \eqref{equ:D}).

%%%% %%%% %%%% %%%% %%%% %%%% %%%% %%%%Gibbs sampling%%% %%%% %%%% %%%% %%%% %%%% %%%% %%%%
%\vspace{1cm}
\subsubsection{MCMC sampling for posterior inference.}
 For applying MCMC
sampling to sample from the posterior, we  look at the conditional distributions. Consider
the conditional distributions at each time point $t$: 
%\vspace{-1mm}
\begin{equation}
\label{conditional}
 \begin{split}
 & p(B_t,A_t| D_t,[B]_{t-},[A]_{t-}, \bullet ) \propto \\	
%& p(B_t,A_t| D_t,[B]_{t-},[A]_{t-}, \alpha,\xi, k ) \propto \\
 & \mathcal{W}_d^-(D_t|\textbf{1},\Delta_t)  P(B_t | B_{t-1})
P(B_{t+1} | B_{t}) 
 P(A_t | A_{t-1}) P(A_{t+1} | A_{t})
\end{split}
%\vspace{-1mm}
\end{equation}

%%%% %%%% %%%% %%%% %%%% %%%% %%%% %%%%posterior sampling B_t%%% %%%% %%%% %%%% %%%% %%%% %%%% %%%%

\paragraph{Posterior sampling for $B_t$.}
The posterior sampling involves sampling assignments. As we are dealing with non-conjugate priors in (\ref{conditional}), 
we use a Gibbs sampling algorithm with $m$ auxiliary variables as presented in \cite{Neal00}. 
%We consider two cases of assignments, the first case considers a truncated Dirichlet process where the number of clusters $k$ is fixed to a large maximal number of clusters $k_{\max}$. 
\changedtext{We consider the infinite model with $k\rightarrow \infty$.}
The aim is to assign one object $l$ in epoch $t$ to either an existing cluster $c$, a new cluster that exists at epoch $t-1$ or epoch $t+1$ or  a totally new cluster. 
The prior probability that object $l$ belongs to an exisiting cluster $c$ at time point $t$ is
 \begin{equation} 
\label{equ:ex_clus}
P(l=c | B_{t-1})P(B_{t+1}|B_t) \propto n_{c_{t-1}}+n_{c_t}^{(-l)})\cdot \frac{n_{c_{t+1}}}{n_{c_t}^{(-l)}} .
\end{equation}

\renewcommand{\arraystretch}{1.5}
\changedtext{There exist four different  prior probabilities of an object $l$ belonging  to  a new cluster $c_{new}$ at time point $t$, which are summarized in table \ref{tab:probs}. }

\begin{table}[h]
\small{
\begin{center}
  \begin{tabular}{ | c  | c c|}
  
\multicolumn{1}{c}{$\boldmath {c_{\textnormal{new}}}$ \bf exists at} & \multicolumn{1}{r}{ \bf$ \bf P(l=c_{new}| B_{t-1})P(B_{t+1}|B_t)$}  \\ \hline
\hline
    
 both time points  $t-1$ and  $t+1$:     & $\propto (n_{c_{t-1}} \cdot \frac{\xi}{m})\cdot n_{c_{t+1}}$& \refstepcounter{equation}(\theequation)\\ \hline
       
time point $t-1$ but not at time point $t+1$: &$ \propto \frac{\xi}{m} \cdot n_{c_{t-1}}$& \refstepcounter{equation}(\theequation)\\ \hline
          
   time point $t+1$ but not at time point $t-1$  &  $\propto \frac{\xi}{m} \cdot n_{c_{t+1}}$&\refstepcounter{equation}(\theequation)  \\ \hline
    
neither   $t-1$ nor  $t+1$,&$\propto \frac{\xi}{m} $ &\refstepcounter{equation}(\label{equ:new_clus}\theequation)\\
 (i.e.\ $l$ belongs to  a completely new cluster) & &\multicolumn{1}{r|}{}\\ \hline
  
     \end{tabular}
  \vspace{0.5cm}
  \caption{\label{tab:probs} \changedtext{Table of prior  probabilities.}}
\end{center}
}
\end{table}

%%%% %%%% %%%% %%%% %%%% %%%% %%%% %%%%MH%%% %%%% %%%% %%%% %%%% %%%% %%%% %%%%

%\vspace{-8mm}
\paragraph{Metropolis-Hastings update steps.}
\label{sec:metro-hast}
In every time point, we need to sample $\beta$ values in the between-class variance matrix $\Sigma_{A_t}$. To find the $\beta$ values within one epoch, we sample the whole ``new" $A_t$ matrix, denoted by $A_{t_{new}}$,
 with a Metropolis-Hastings algorithm (see \cite{RobertCasella}). 
With $A_{t_{old}}$ we denote the initial $A_t$ matrix.
\changedtext{ As \emph{proposal distribution} we chose a Wishart distribution, and for the \emph{prior} we chose a Wishart distribution as well,
leading to  $P(A_{t_{new}}|A_{t_{old}})\sim \mathcal{W}(A_{t_{new}}|A_{t_{old}})$ and
       $P(A_{t_{new}})\sim \mathcal{W}(A_{t_{new}}|I_{k_{b_t}}).$}
%       This leads to:
%\begin{eqnarray*}
%&&\theta<\frac{\mathcal{W}_d^-(D|1,\Delta_{B_{new}})\mathcal{W}(A_{new}|I_k)}{\mathcal{W}^-_d(D|1,\Delta_B)\mathcal{W}(A_{old}|I_k)}
%\frac{\mathcal{W}(A_{old}|A_{new})}{\mathcal{W}(A_{new}|A_{old})}\\
%\end{eqnarray*}

 %As the proposal distribution we choose a Wishart distribution.

\paragraph{Hyperparameters and Initialization.}
\changedtext{Our model includes the following hyperparameters: the scale parameter  $\alpha$, the number $k$ of clusters, the Dirichlet rate $\xi$, the degrees of freedom $d$ and a scale parameter $s$. The model is not sensitive to the choice of $s$, and we fix $s$ to 1. $\alpha$ is sampled from a Gamma distribution with  shape and scale parameters $r$ and $m$.  For the number $k$ of clusters, our framework is applicable to two scenarios: we can either assume $k=\infty$ which results in the CRP model, or we fix $k$ to a large constant which can be viewed as a truncated Ewens process. As the model does not suffer from the label switching problem, initialization is not a crucial problem.  We initialize  the block size with size 1, \ie we start with one cluster for all objects. The
Dirichlet rate $\xi $ only weakly influences the likelihood, and the variance only
decays with $1/\log(n_t)$ (see \cite{Ewens72}). In practice, we
should not expect to reliably estimate $\xi$. Rather,
we should have some intuition about $\xi$, maybe guided by
the observation that under the Ewens process model the
probability of two objects belonging to the same cluster is $1/(1+\xi)$. We can then either define an appropriate prior
distribution, or we can fix $\xi$. Due to the weak effect of $\xi$ on
conditionals, these approaches are usually very similar.
The degrees of freedom $d$ can  be estimated by the rank of $K$, if it is known from a pre-processing procedure. As $d$ is not a very critical parameter (all likelihood contributions are basically raised to the power of $d$), $d$
might also be used as an annealing-type parameter for freezing a representative partition in the limit for $d\rightarrow \infty$. }

\paragraph{Pseudocode.}
A pseudocode of the sampling algorithm is given in Algorithm 1.
\vspace{-3mm}
\begin{algorithm}[h]

\begin{algorithmic}

\FOR{ $i=1$ to iteration}
\FOR{ $t=1$ to $T$}
\FOR{ $j=1$ to $n_t$}
\STATE Assign one object to an existing cluster or a new one using Eqs.~\eqref{conditional}-\eqref{equ:new_clus}\\
 Update  $k_{b_t}$\\
 \ENDFOR
 \ENDFOR
 
 \FOR{ $t=1$ to $T$}
\STATE Sample new $A_t$ matrix using Metropolis-Hastings 
\ENDFOR
%\STATE resample $\alpha$ from a Gamma distribution.
\ENDFOR

\end{algorithmic}
\caption{\label{algo1}  Pseudocode Te-TiWD}
\end{algorithm}

\vspace{-3mm}
\paragraph{Complexity.}
\changedtext{We define  one sweep of the Gibbs sampler as one complete update of $(B_t, A_t)$. The most time consuming part in a sweep is the update of $B_t$ by re-estimating the assignments to blocks for a single object (characterized by a row/column in $D_t$), given the partition of the remaining objects. Therefore we have to compute the membership probabilities in all existing blocks (and in a new block). Every time a new partition is analyzed, a naive implementation requires $O(n^3)$ costs for computing the determinant of $\tilde{W}_t$ and the product $\tilde{W}_tD_t$. In one sweep we need to compute $k_{b_t}$ such probabilities for $n_t$ objects, summing up to costs of $O(n^4k_{b_t})$.  
This suggests that the scalability to large datasets can pose a problem. In this regard we plan to address run time in future work by investigating the potential of variational methods, parallelizing the MCMC sampler  and by updating parameters associated with multiple time points simultaneously.
}

%%%% %%%% %%%% %%%% %%%% %%%% %%%% %%%%Identifiability%%% %%%% %%%% %%%% %%%% %%%% %%%% %%%%

\paragraph{Identifiability of clusters.} In some applications, it is of interest to identify and track clusters over time. For example by grouping newspaper articles into topics it might be interesting  to know which topics are present over a long time period, when a new topic becomes popular and when a former popular topic dies out.  Due to the translation-invariance of our novel longitudinal model, we additionally need a cluster mean to be able to track  clusters over the time course. To estimate the mean of the clusters we propose to embed the ``overall" data matrix $D^*\in \R^{N \times N}$ with $N:=\sum_{t=1}^T n_t$  that contains the pairwise distances between all objects over all time points into a  vector space, using kernel PCA. We first construct a positive semi-definite matrix $K^*$  which fulfills $D^*_{ij}=K^*_{ii}+K^*_{jj}-2K^*_{ij}$. For correcting $K^*$, we compute the centered matrix $K^*_c=Q^*K^*Q^*$ with $Q^*_{ij}=\delta_{ij}-\frac{1}{N}$. As a next step, we compute the eigenvalue decomposition of $K^*_c$, i.e.\ $K^*_c=V\Lambda V^T$ and then project on the principal axes $X^*=V\Lambda^{\frac{1}{2}}$, i.e.\ we use the principal components as coordinate axes.
 By embedding the distances $D^*$ into a vector space, the  underlying block structure might be distorted (see Fig.~\ref{fig:information_loss}).  As our aim is to find the underlying block structure, it is hence infeasible to embed the data for clustering. But, for tracking the clusters, we just need to find the mean of an already inferred block structure, i.e.\ we embed the data not for grouping  data points, but for finding a mean of an already assigned partition that allows us to track the clusters over time. We embed all objects together  and choose the same orthogonal transformations for all objects, which enables identifiability of cluster means over the time course. This preprocessing step is only necessary if one is interested in the identifiability of clusters, and $X^*$ needs only to be computed once outside the sampling routine. \changedtext{Since computing $X^*$ is computationally expensive, it is done only once as a preprocessing step if required. Computing $X^*$ within the sampling routine would slow down our sampler significantly.} 

%%%%%%%%%%%%%%%%%%%% %%%% %%%%Experiments%%%%%%%%%%%%%%%%% %%%% %%%%

\section{Experiments}
\label{experiments}

\subsection{Synthetic Experiments}
\subsubsection{Well separated clusters.}
\begin{figure}[h!]
  \begin{center}
    \vspace*{-4ex}
    \begin{minipage}{0.13\linewidth}
      \centering
      \includegraphics[height=3.5cm]{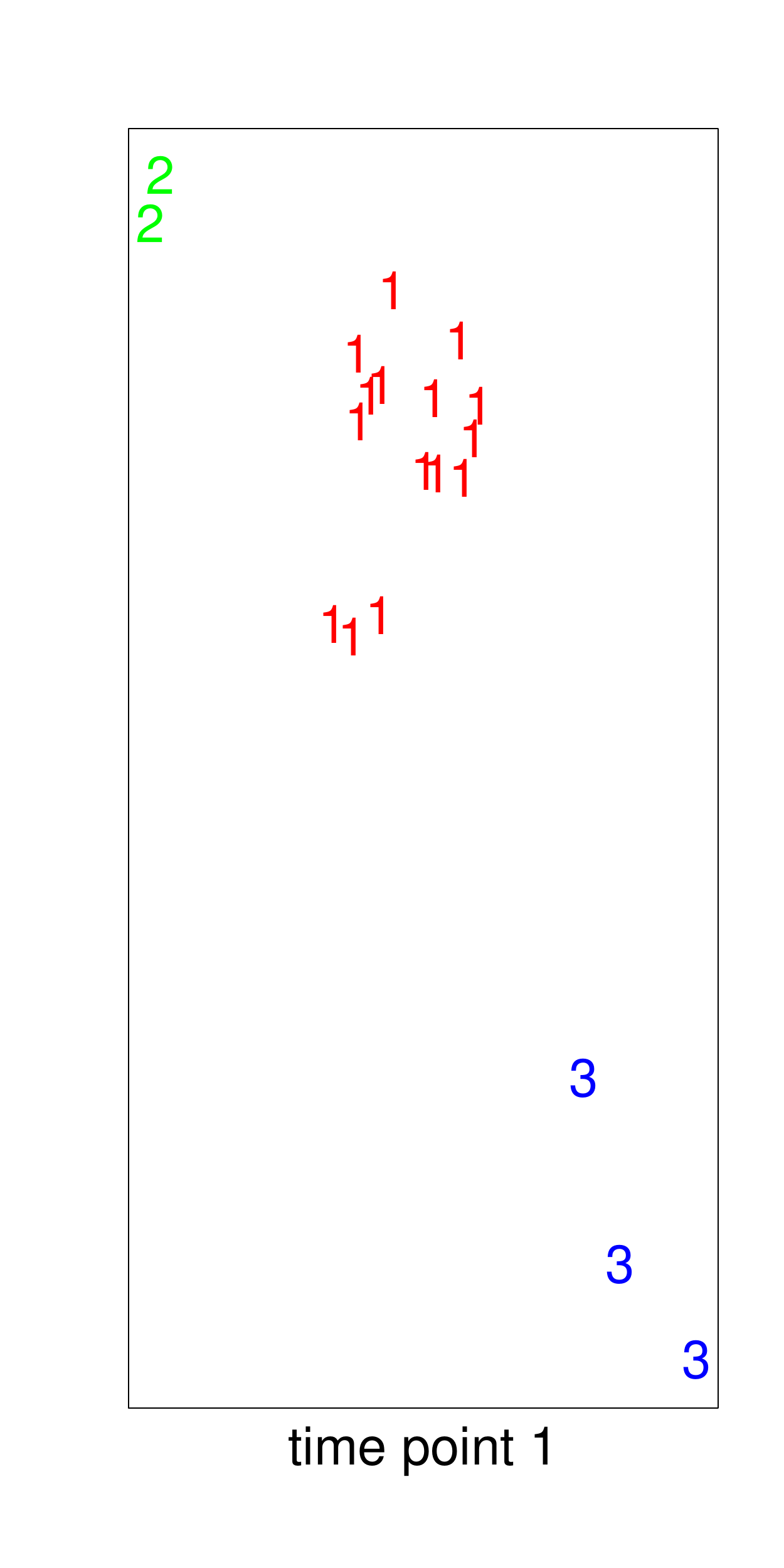}
    \end{minipage}
    $\xrightarrow{\textnormal{time}}$
    \begin{minipage}{0.13\linewidth}
      \hspace{-3mm}
      \centering
      \includegraphics[height=3.5cm]{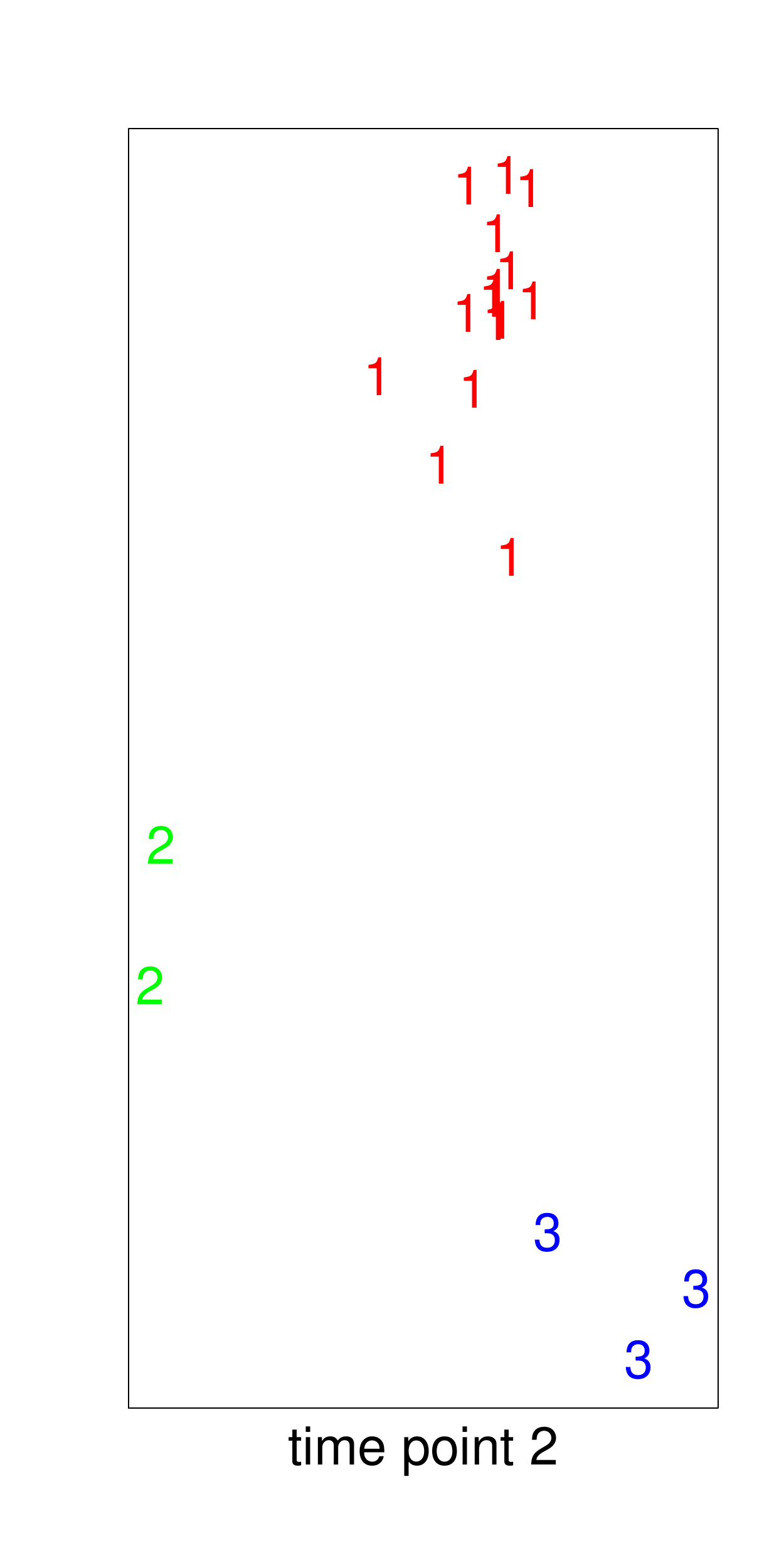}
    \end{minipage}
    $\xrightarrow{\textnormal{time}}$
    \begin{minipage}{0.13\linewidth}
      \hspace{-3mm}
      \centering
      \includegraphics[height=3.5cm]{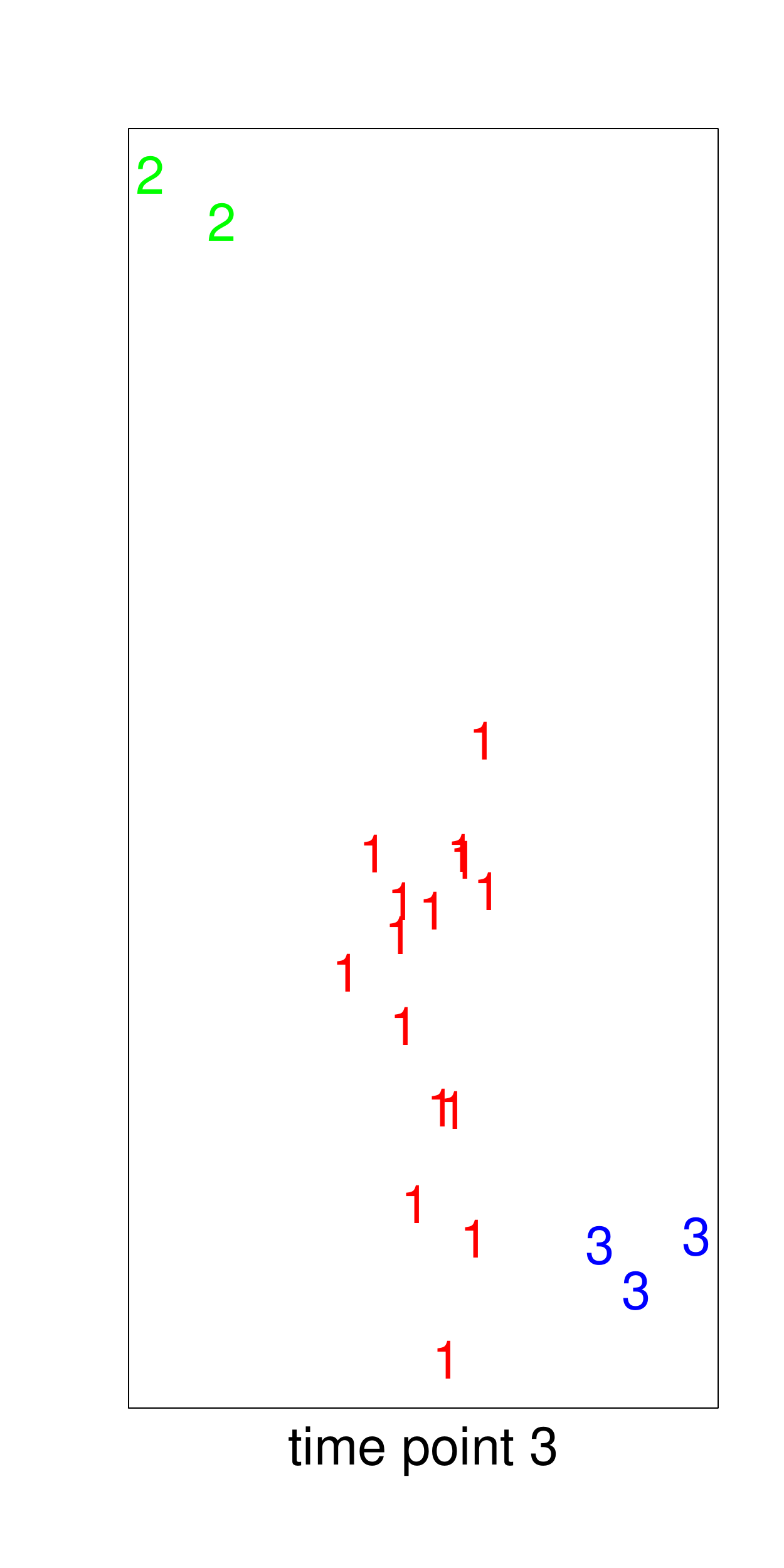}
    \end{minipage}
    $\xrightarrow{\textnormal{time}}$
    \begin{minipage}{0.13\linewidth}
      \hspace{-3mm}
      \centering
      \includegraphics[height=3.5cm]{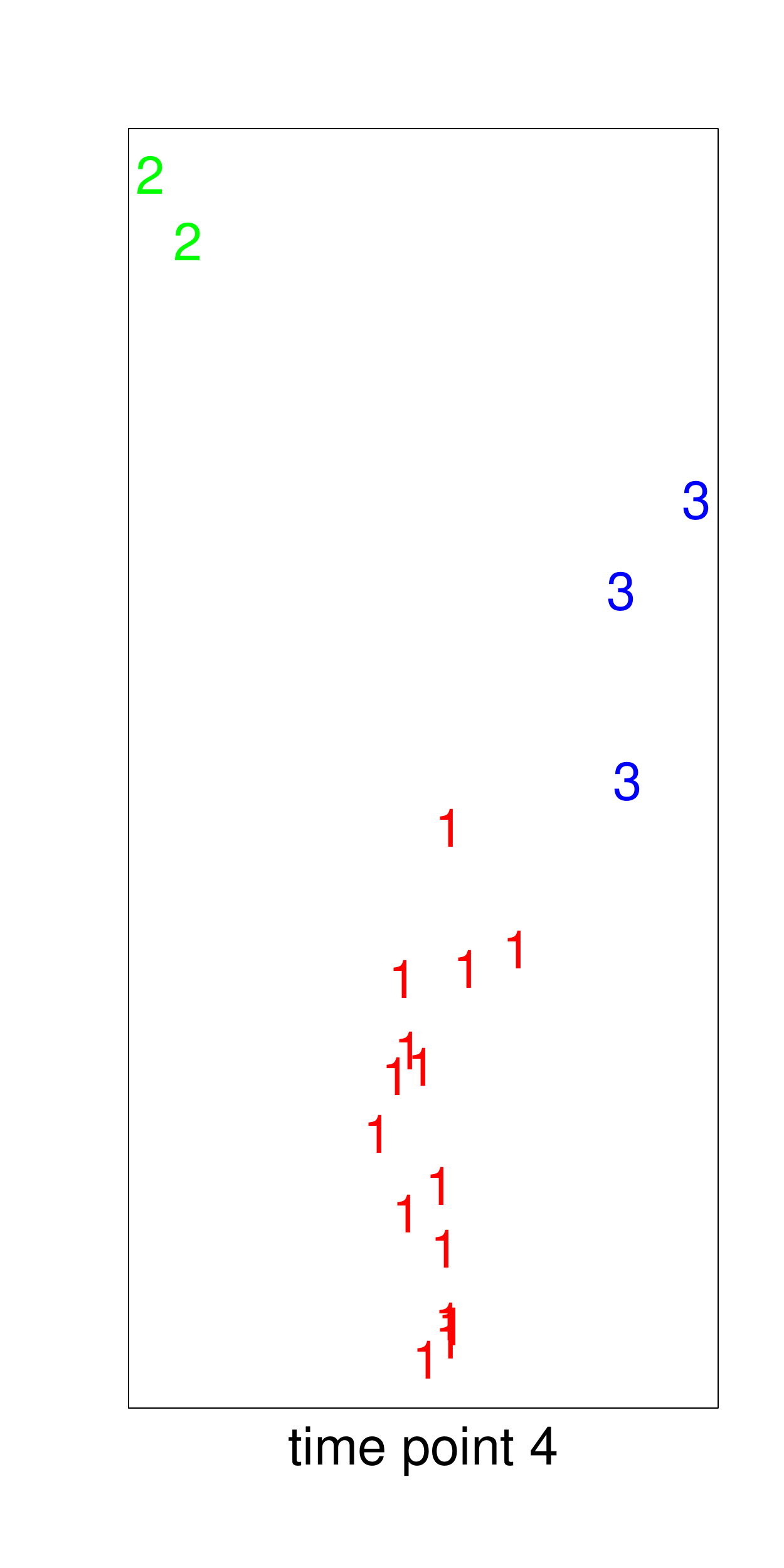}
    \end{minipage}
    $\xrightarrow{\textnormal{time}}$
    \begin{minipage}{0.13\linewidth}
      \hspace{-3mm}
      \includegraphics[height=3.5cm]{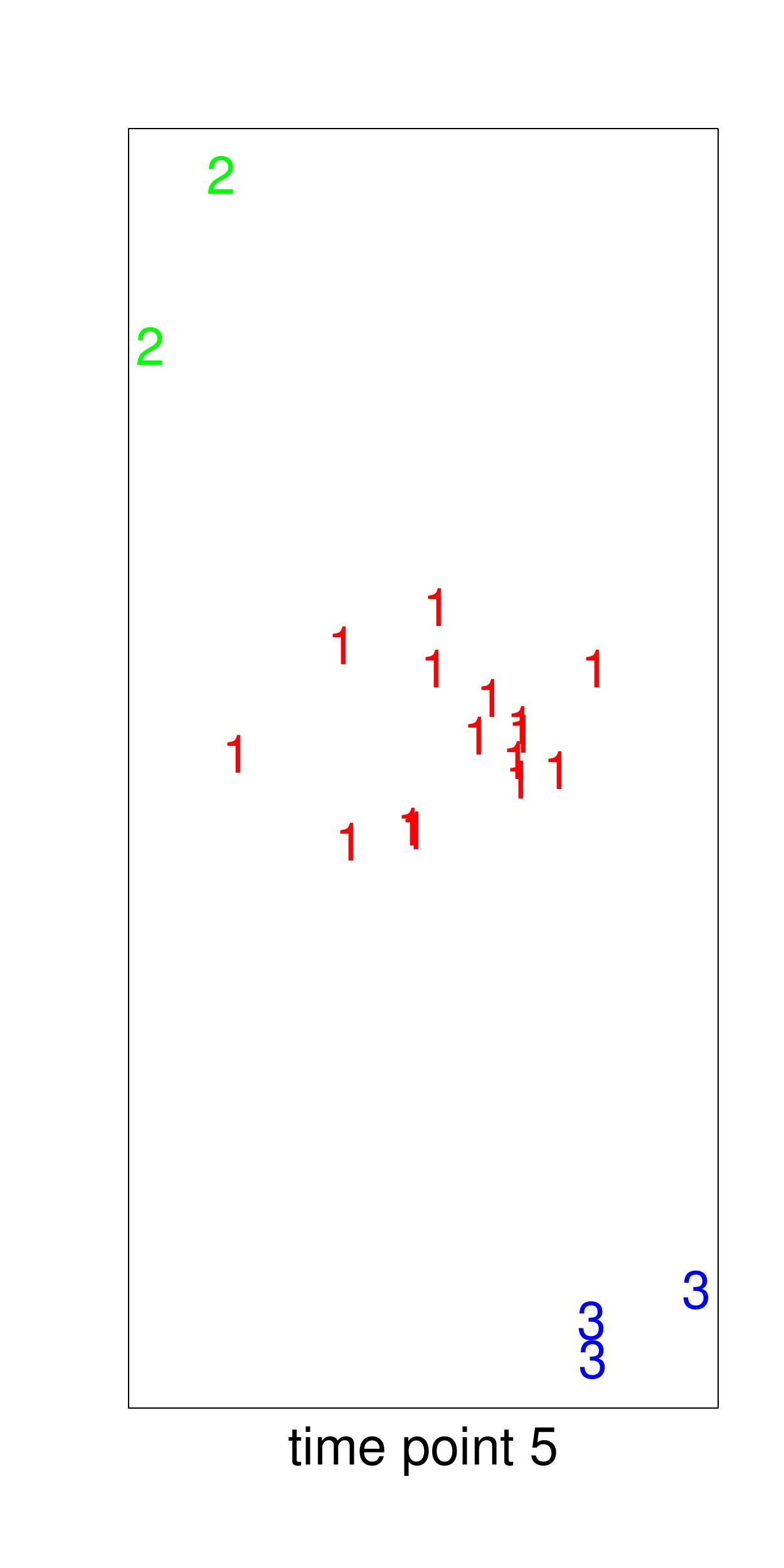}
    \end{minipage}
  \end{center}
  \caption{\label{1_exp_pca} PCA projections of five time points with
    three well separated clusters per time point. Numbers and colors
    correspond to true labels.\vspace*{-4ex}}
\end{figure}

In a first experiment, we test our method on simulated data. \changedtext{We simulate data in two ways, first we generate data points accordingly to the model assumptions, and secondly we generate data independent of the model assumptions. }We start
with a small experiment where we consider five time points each with
20 data points per time point in 100 dimensions, \ie we consider a
small data set size and large dimension problem.
\vspace{-5mm}
\paragraph{Data generation.} \changedtext{The data  is generated (according to the model assumptions) in the following way: 
for the first time point, a random block matrix $B_1$ of size $n_1=20$
is sampled with $k_{b_1}=3$ (i.e.~we generate 3 blocks at time point 1)}. A $k_{b_1}\times k_{b_1}$
matrix $A_1$ is sampled from $\mathcal{W}_d(I_{k_{b_1}})$ 
and $B_1$ is filled with the corresponding $\beta$ values from $A_1$,
which leads to the $n_1\times n_1$ matrix
$\Sigma_{K_{1}}$.  Next,
$d_1=100$ samples from $\mathcal{N}(0_n,\Sigma_{B_1})$ are drawn with
$\Sigma_{B_1}=\alpha I_{n_1}+\Sigma_{A_{1}}$, where $\alpha=2$, and
stored in the $(n_1 \times d_1)$ matrix $X_1$.  By choosing $\alpha=2$, we create well separated clusters. The similarity matrix
$K_1=X_1X_1^T$ is computed and squared distances are stored in matrix
$D_1$.  For the following time points $t>1$, the partition for the
block matrix $B_t$ of size $n_t$ is drawn from a Dirichlet-Multinomial
distribution, conditioned on the partition at time point $t-1$.  A new
$A_t$ matrix is sampled from $\mathcal{W}_d(A_{t-1})$. If the number
of blocks in time points $t$ and $t-1$ are different, we sample $A_t$
according to Eq.~\eqref{newrow}.  $d_t$ samples from
$\mathcal{N}(0_{n_t},\Sigma_{B_t})$ are drawn with
$\Sigma_{B_t}=\alpha I_{n_t}+\Sigma_{A_t}$. The pairwise distances are
stored in the matrix $D_t$.  A PCA projection of this data is shown in
Fig.~\ref{1_exp_pca} for illustration.

\begin{figure}[h]
    \begin{center}
      \vspace*{-6ex}
      \includegraphics[height=5.5cm]{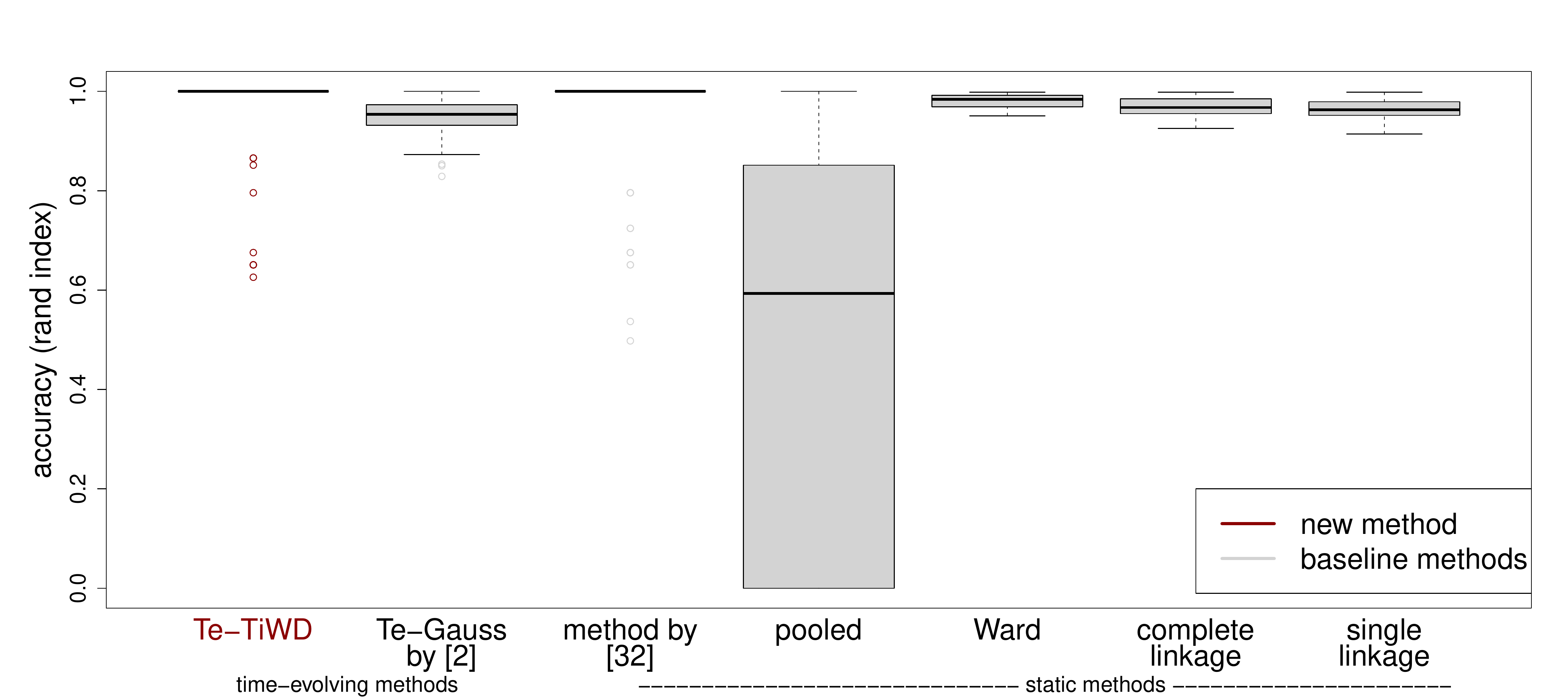}
    \end{center}
    \caption{\label{1_exp_plots} We compare our new dynamic model
      (Te-TiWD) with baseline methods: static clustering as in
      \cite{Vogt2010}, combined clustering over all time points (pooled),
      a Gaussian time-evolving clustering model (Te-Gauss) as well as to
      Ward, complete linkage and single linkage. In this experiment with
      three well separated clusters per time point, all methods perform
      very well, except for pooling the data.}
  \end{figure}

\paragraph{Experiments.} \changedtext{We perform four illustrative experiments for well-separated data:}
\begin{enumerate}
\item[a)] 500 Gibbs sweeps are computed for the Te-TiWD cluster process
  (after a burn-in phase of 250 sweeps). \changedtext{We check
  convergence of the algorithm by analyzing the trace of the number of
  blocks $k_{b_t}$ during sampling.  On this trace plot we observe
  after how many sweeps the sampler stabilizes (the number of sweeps
  depends on the size of the data set). We observe a remarkable
  stability of the sampler (compared to the usual situations in
  traditional mixture models), which follows from the fact that no
  label-switching can appear.  Finally, we perform an annealing
  procedure to \emph{freeze} a certain partition. Here, $d$ is used as
  an annealing-type parameter for freezing a representative partition
  in the limit $d\rightarrow\infty$. On a standard computer, this
  experiment took roughly 4 minutes, and the sampler stabilizes after
  roughly 50 sweeps.}  As the ground truth is known, we can compute
  the adjusted rand index as an indicator for the accuracy of the
  Te-TiWD model.  We repeat the clustering process 50 times. The
  result is shown in form of a box plot (Te-TIWD) in
  Fig.~\ref{1_exp_plots}.
  
\item[b)] In order to compare the performance of the time-evolving
  model (Te-TiWD) to baseline models, we also run the static
  probabilistic clustering process as well as hierarchical clustering
  models (Ward, complete linkage and single linkage) on every time
  point separately and compute the averaged accuracy over all time
  points. For the comparison to the static probabilistic method
  \cite{Vogt2010}, we use the same set-up as for Te-TiWD, we run 500
  Gibbs sweeps with a burn-in phase of 250 sweeps and repeat it for 50
  times.  For the hierarchical methods, the resulting trees are cut at
  the number of clusters found by the nonparametric probabilistic
  model. Accuracy is computed for every time point separately, and
  then averaged over all time points. In this scenario, the static
  clustering models performs almost as well as the time-evolving
  clustering, see Fig.~\ref{1_exp_plots}, as expected in such a
  setting where all groups are well separated at every single time
  point.

\item[c)] As a further comparison to a baseline dynamic clustering model,
  we embed the distances into a Euclidean vector space and run a
  Gaussian dynamic clustering model (Te-Gauss) on the embedded
  vectorial data. As the clusters are well separated, embedding the
  data and clustering on vectors works well, as shown in box plot
  ``Te-Gauss" in Fig.~\ref{1_exp_plots}.

\item[d)] As a last comparison we evaluate a pooled clustering over all
  time points. For this experiment, we not only need the pairwise
  distances at every single time point, but also the pairwise
  distances of objects across all time points.  The number of sweeps
  and repetitions remains the same as in the experiments above.  We
  conduct one clustering over all objects of all time points, and
  after clustering, we extract the objects belonging to the same time
  point and compute the rand index on every time point
  separately. This experiment shows worse results (see box plot
  ``pooled" in Fig.~\ref{1_exp_plots}), which can be explained as
  follows: by combining all time points to one data matrix, new
  clusters over all time points are found, this means clusters are
  shifted and objects over time are grouped together, introducing new
  clusters by reforming boundaries of old clusters. These new clusters
  inhibit objects to group together which would group together at
  single time points, destroying the underlying ``true" cluster
  structure.
\end{enumerate}

\subsubsection{Highly overlapping clusters.}
\begin{figure}[h!]
    \centering
    \includegraphics[height=6cm]{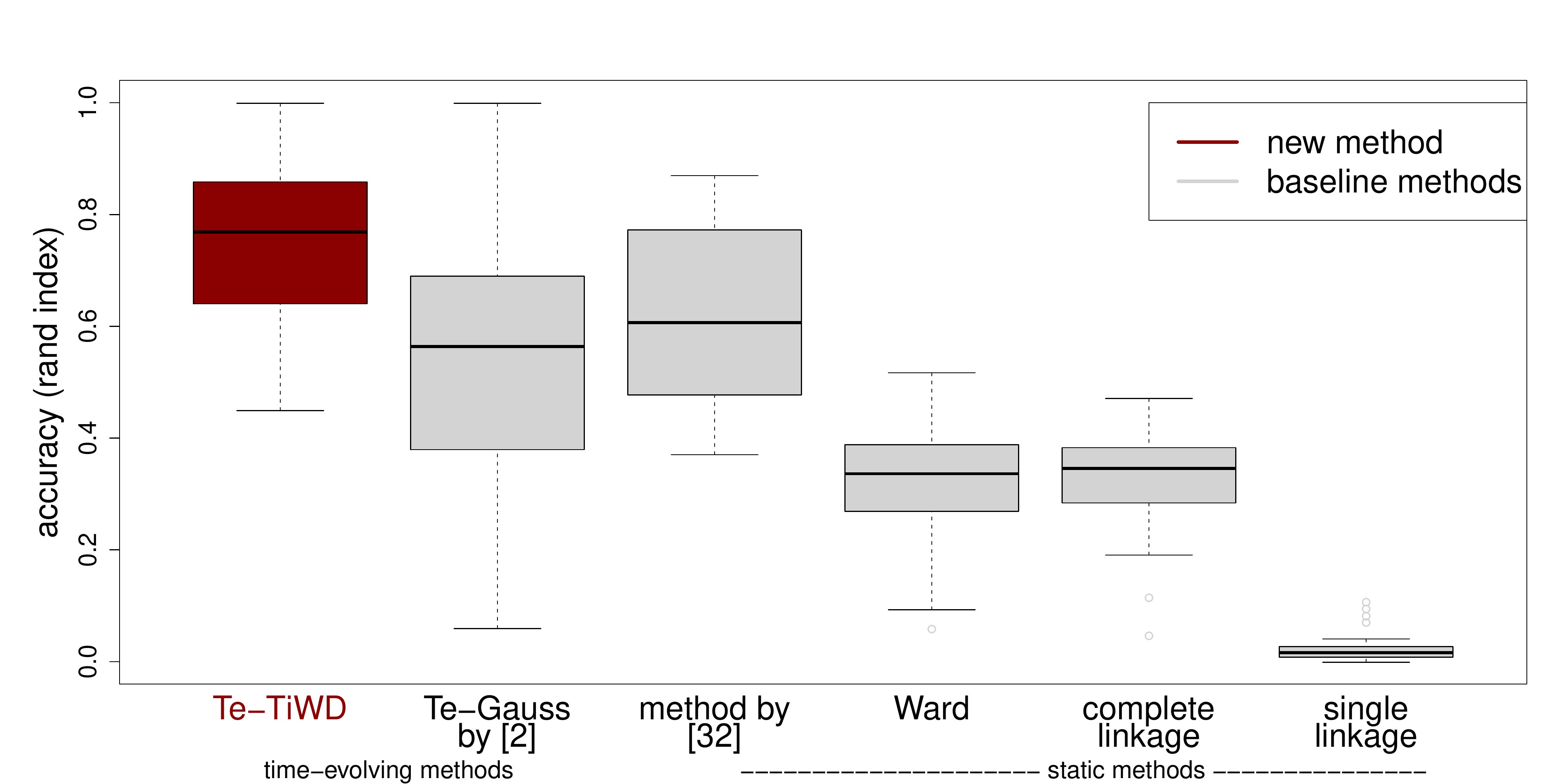} 
    \caption{\label{2_exp_plots} We compare our new model (Te-TiWD)
      with baseline methods on synthetic data for five highly
      overlapping clusters. Our model significantly outperforms all
      baseline methods.\vspace*{-5ex}}
  \end{figure}

\begin{figure}[b!]
    \begin{center}
      \vspace*{-4ex}
      %\hspace{-5mm}
      \begin{minipage}{0.13\linewidth}
        \centering
        \includegraphics[height=3.5cm]{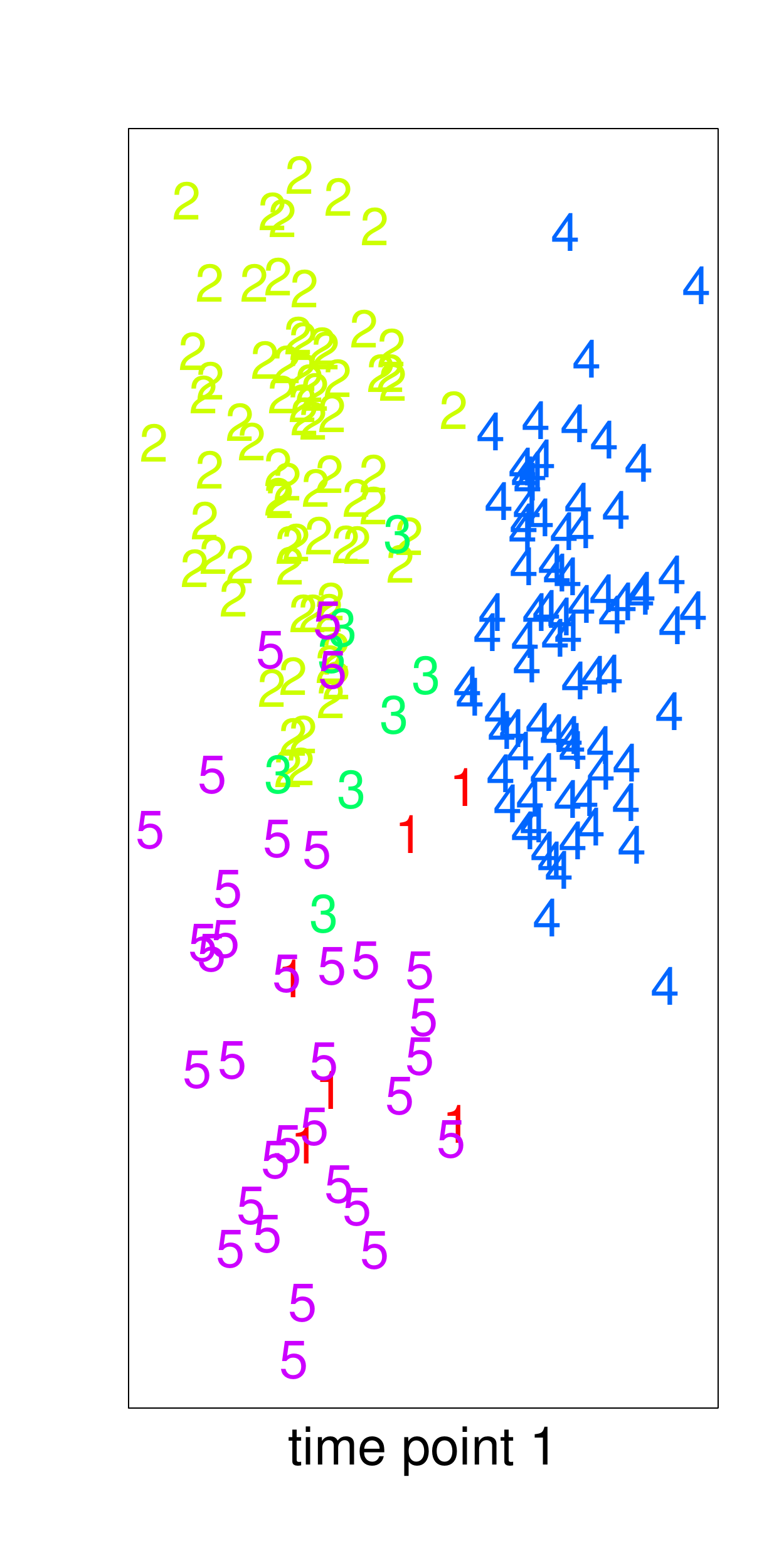}
      \end{minipage}
      $\xrightarrow{\textnormal{time}}$
      \begin{minipage}{0.13\linewidth}
        \hspace{-3mm}
        \centering
        \includegraphics[height=3.5cm]{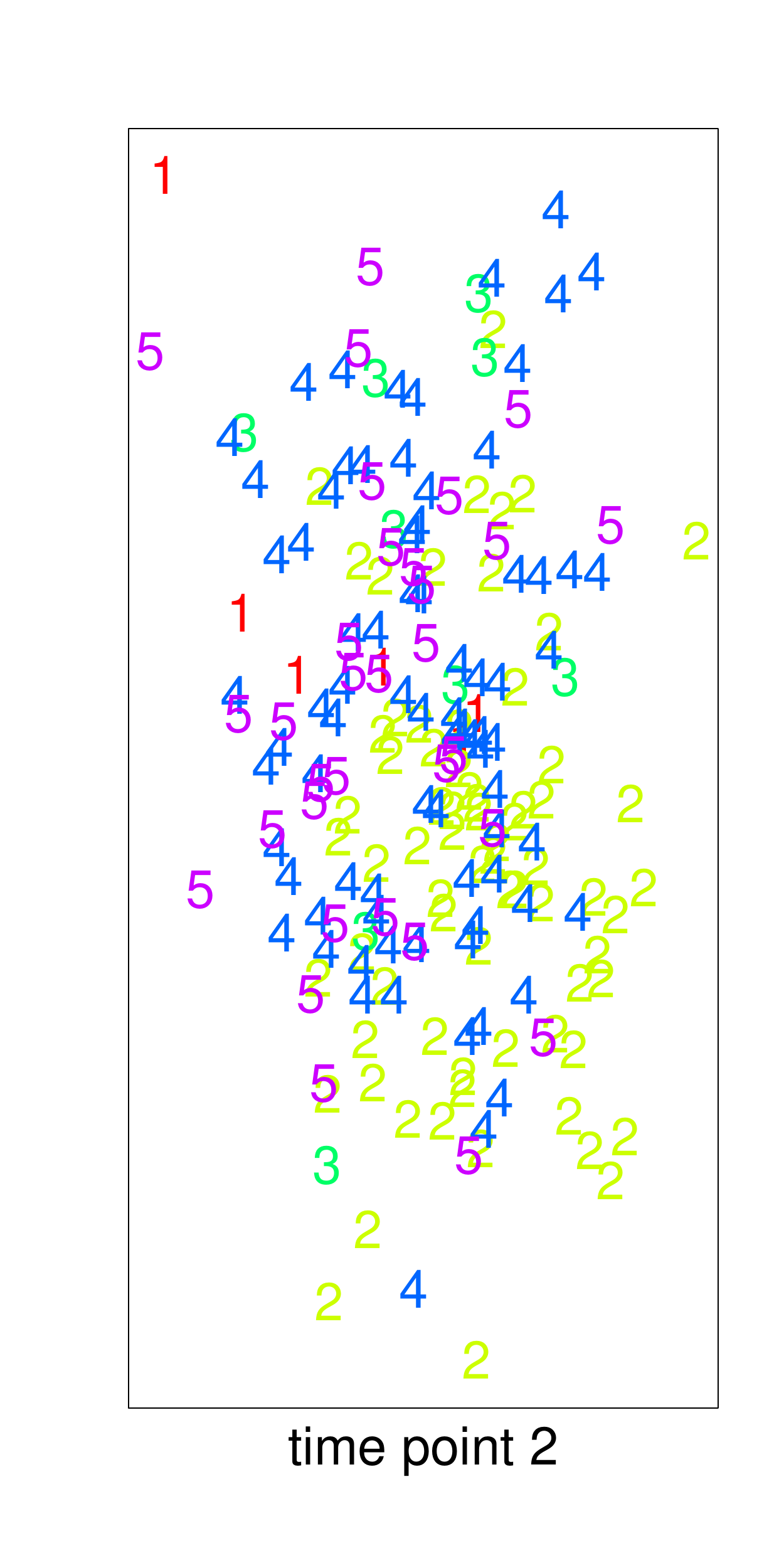}
      \end{minipage}
      $\xrightarrow{\textnormal{time}}$
      \begin{minipage}{0.13\linewidth}
        \hspace{-3mm}
        \centering
        \includegraphics[height=3.5cm]{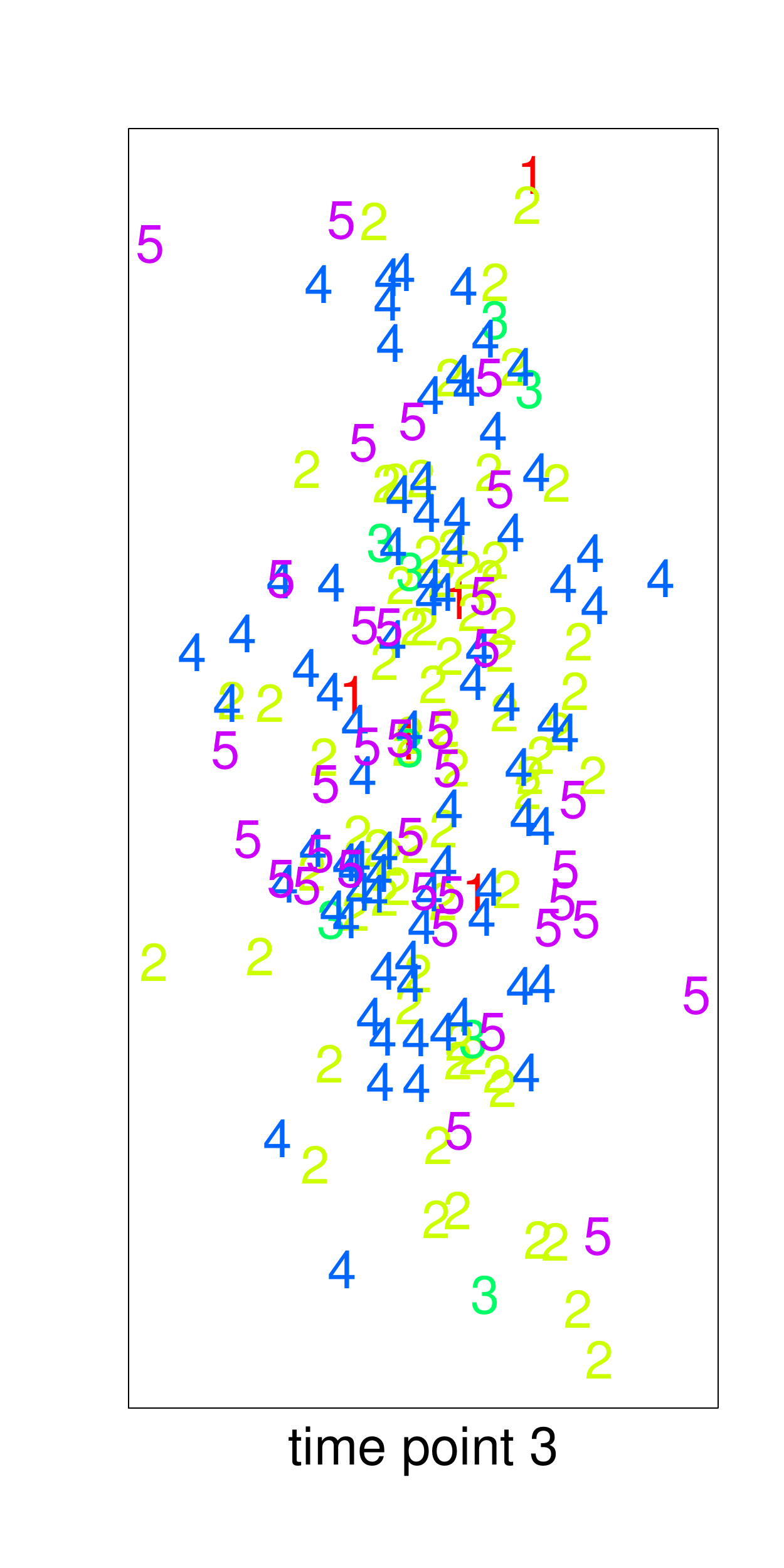}
      \end{minipage}
      $\xrightarrow{\textnormal{time}}$
      \begin{minipage}{0.13\linewidth}
        \hspace{-3mm}
        \centering
        \includegraphics[height=3.5cm]{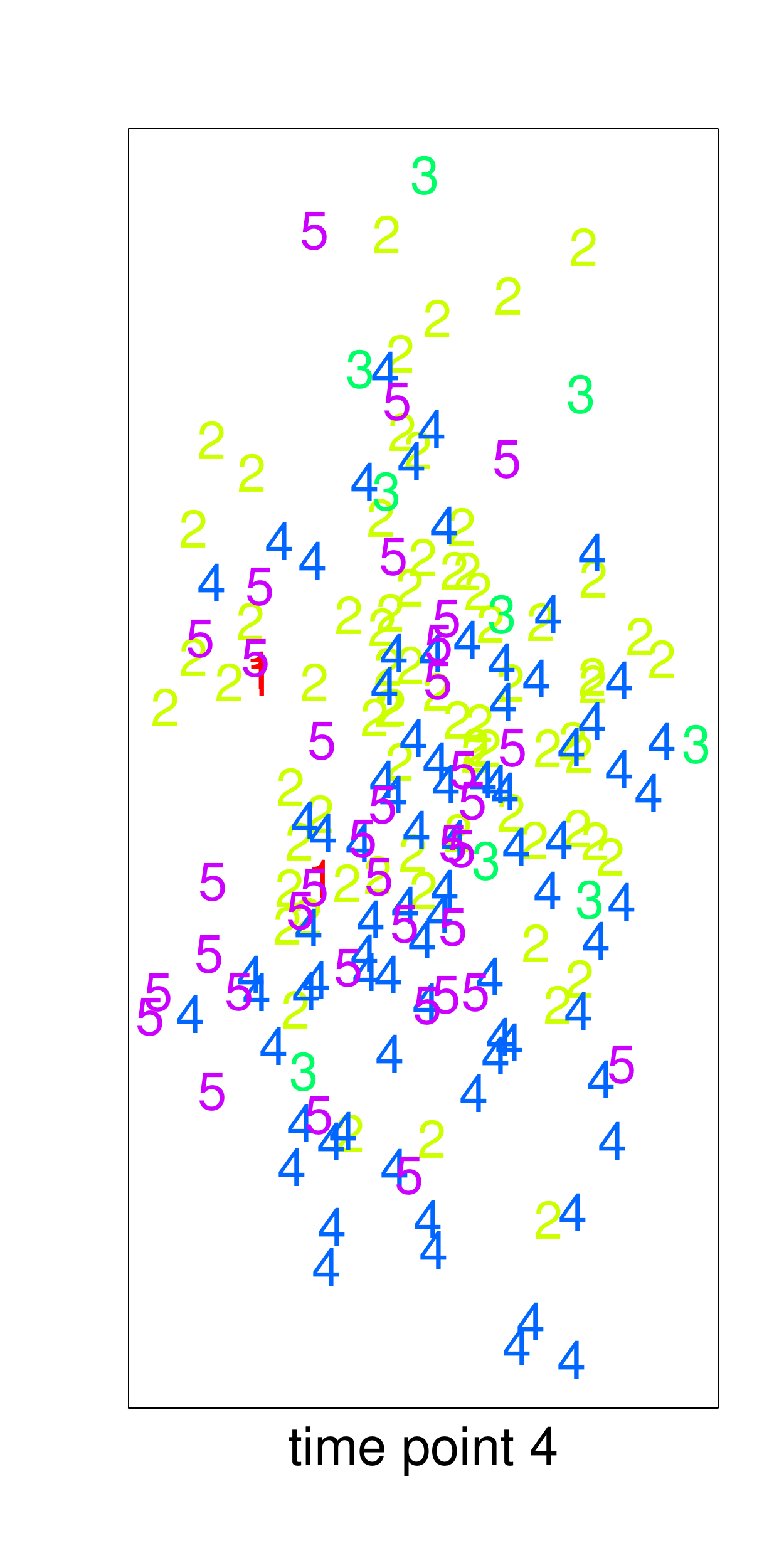}
      \end{minipage}
      $\xrightarrow{\textnormal{time}}$
      \begin{minipage}{0.13\linewidth}
        \hspace{-3mm}
        \centering
        \includegraphics[height=3.5cm]{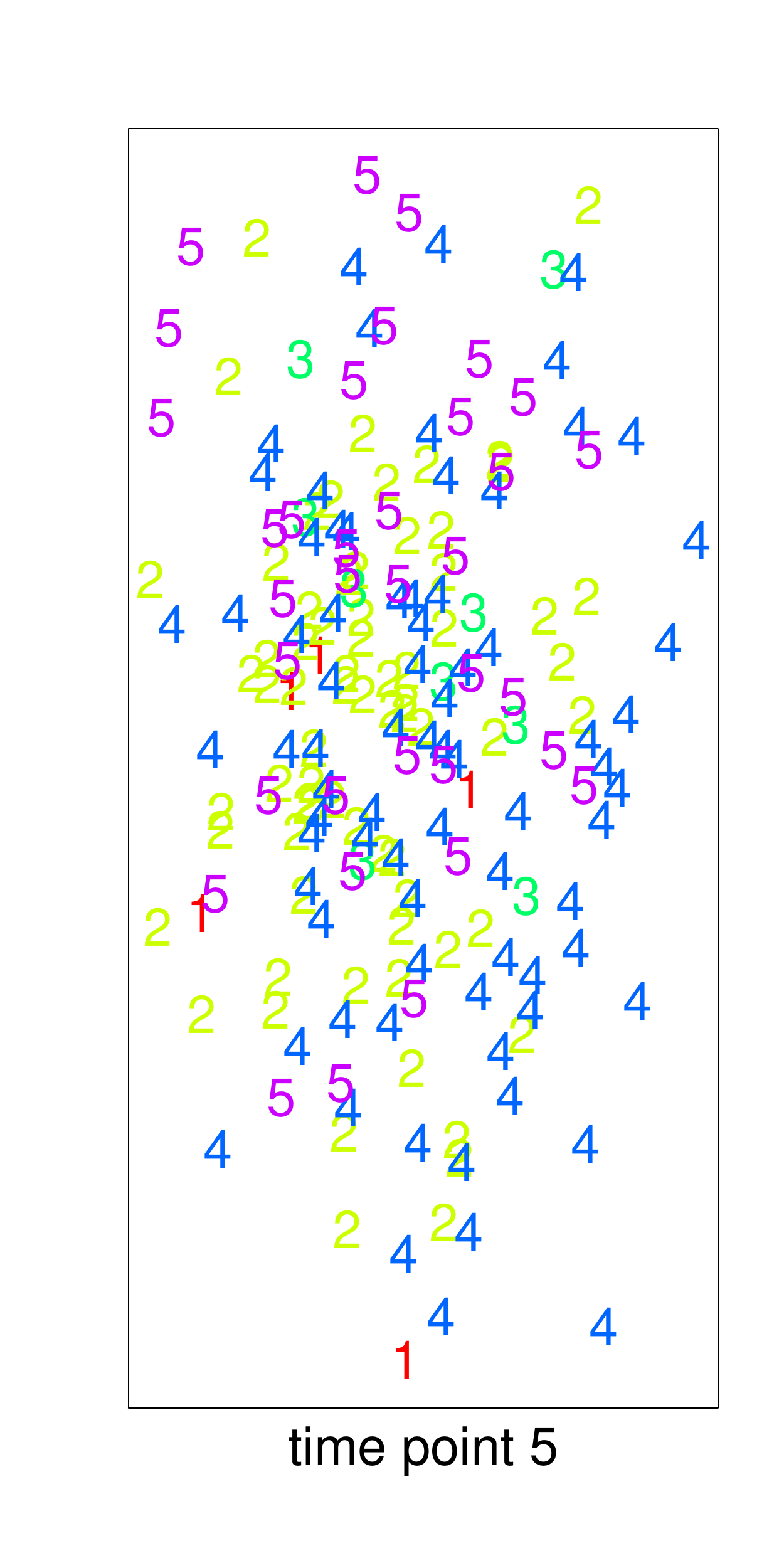}
      \end{minipage}
    \end{center}
    \caption{\label{2_exp_pca}  PCA projections of  five time points of simulated data with five highly overlapping clusters. Numbers and colors correspond to true labels.}  
  \end{figure}

  For a second experiment, we generated data in a similar way as
  above, but this time we create 5 highly overlapping clusters each
  with 200 data points per time point in 40 dimensions. A PCA
  projection of this data is shown in Fig.~\ref{2_exp_pca}. On a standard computer, this
  experiment took roughly 3  hours, and the sampler stabilizes after
  roughly 500 sweeps. Again, we
  compare the performance of the translation-invariant time-evolving
  clustering model with static state-of-the-art probabilistic and
  hierarchical clustering models which cluster on every time point
  separately and a time-varying Gaussian clustering model on embedded
  data (Te-Gauss).  For highly overlapping clusters, the new dynamic
  clustering model outperforms the static probabilistic clustering
  model \cite{Vogt2010}, and the hierarchical models (Ward, complete
  linkage, single linkage) fail completely.  Further, our new model
  Te-TiWD outperforms the dynamic, vectorial clustering model
  (Te-Gauss), demonstrating that embedding the data into a Euclidean
  vector space yields worse results than working on the distances
  directly. \changedtext{We tested the statistical significance with
    the Kruskal-Wallis rank-sum test and the Dunn post test with
    Bonferroni correction for pairwise analysis. These tests show that
    Te-TiWD performs significantly better than all clustering models
    we compared to}. Results are shown in Fig.~\ref{2_exp_plots}.

%\vspace{-8mm}
\subsubsection{Data generation independent of model assumptions.}

\vspace{-0.3cm}
  \begin{figure}[h!]
  \centering
\includegraphics[height=6cm]{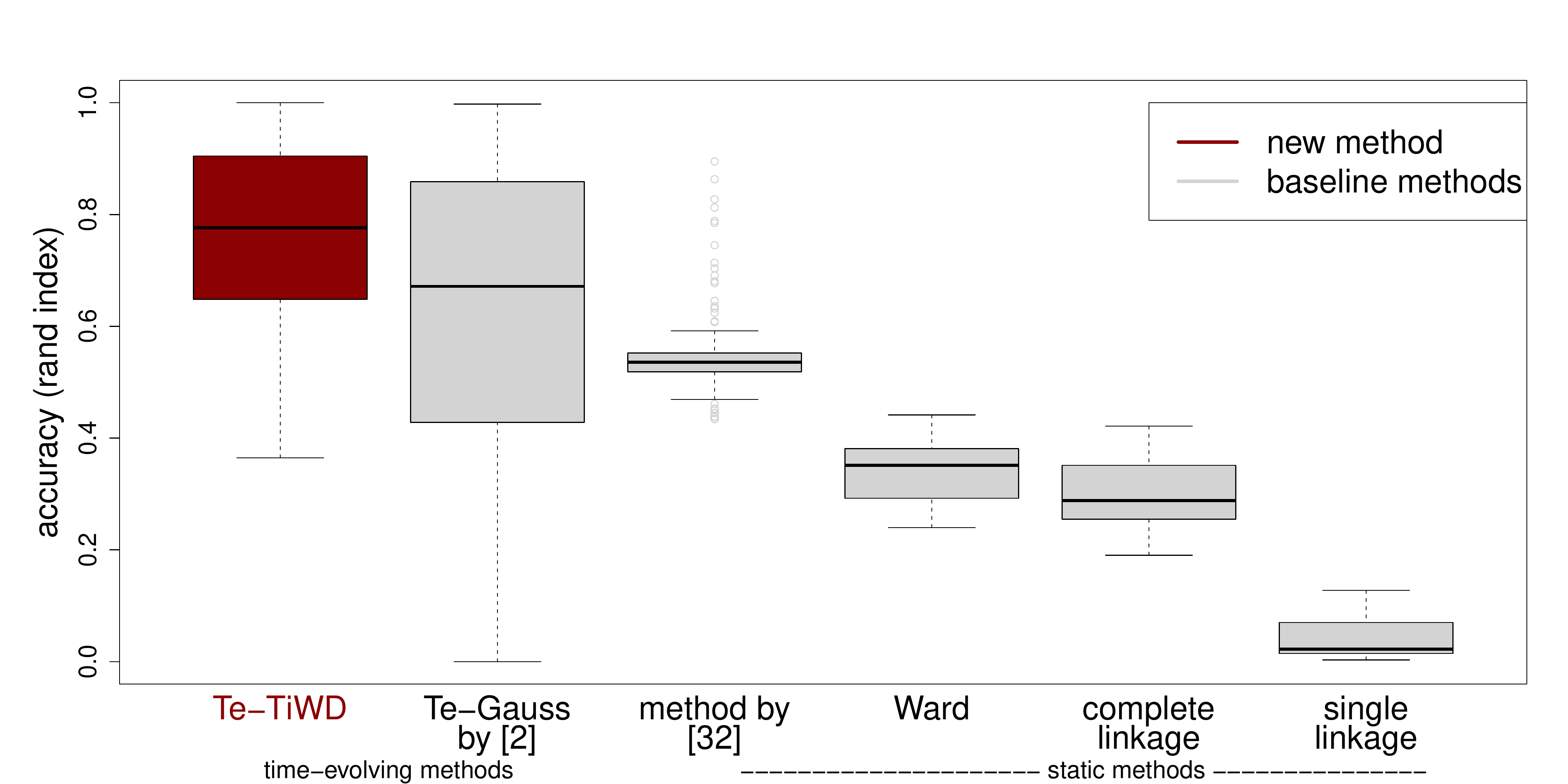} 
 \caption{\label{3_exp_plots}  %3 well separated clusters with 20 data points in 100 dimensions. %Experiment was done  at 5 time-points, a projection of one time point is shown in this figure. 
 We compare our new  model (Te-TiWD) with baseline methods on synthetic data which is generated independent of the model assumptions for five highly overlapping clusters. We observe that our method significantly outperforms all baseline methods.  }
\end{figure}
We also generate data in a second way which is independent of the model assumptions to demonstrate that the performance of our model Te-TiWD is independent of the way  the data was generated. To demonstrate  this, we repeat the case of highly overlapping clusters over 5 time points and  generate data in the following way:
dynamic Gaussian clusters are generated over a period of 5 time points. At each time point five clusters are generated.  200 data points are available at every time point and randomly split into 5 parts, every part representing the number of data points per cluster. For consecutive time points, the number of data points per cluster is sampled  from  a Dirichlet-Multinomial distribution. Every cluster is sampled from a Gaussian distribution with a large variance, resulting in highly overlapping clusters.  Between time steps, the cluster centers move randomly, with relocations sampled from the same distribution. Finally, at every time point, the model-based pairwise distance matrix $D_t$ is computed, resulting in a series of moving distance matrices.
 On this second synthetic data set, Te-TiWD performs  significantly better than all baseline methods as well, as shown in Fig.~\ref{3_exp_plots}.   Note that for the comparison with the Gaussian dynamic clustering model (Te-Gauss) we first embed the distances $D_t$ into vectorial data $X_t^*$ and do not work on the simulated vectorial data directly, to obtain a fair comparison.
% A PCA projection  this data is also shown in Fig.~\ref{3_exp_pca}.

%%%%%%%%%%%%%%%%EHR DATAl%%%%%%%%%%%%%
%\vspace{-1cm}
\subsection{Analysis of Brain Cancer Patient based on Electronic Health Records (EHR)}
\label{sec:EHR}
\changedtext{We apply our proposed model to a dataset of clinical
  notes from brain cancer patients at Memorial Sloan Kettering Cancer
  Center (MSKCC). Brain cancer patients make up 1.4\% of all cancer
  patients, annually. Survival is highly variable, depending on age, 
  gender, cancer subtype, and progression when caught, but
   on average 33\% of patients survive the first five years.\\[0.5ex]
As a first step, we partition a total of 195,297 sentences from 3,403
electronic health records (EHR) from 704 MSKCC brain cancer patients
into groups of similar vocabulary.  This is done by treating sentences
as binary vectors with non-zero entries corresponding to vocabulary,
and obtaining a similarity measure using ranked neighborhood
comparisons \cite{RANC}.  Sentences are clustered using this
similarity measure with the Louvain method \cite{louvain}.  Using
these sentence clusters as features, we obtain patient similarities
with the same ranked neighborhood comparison method.  We partition the
patients documents into windows of one year each, and obtain three
time points where enough documents are available to compute
similarities between patients.  At each year, we represent a patient
with a binary vector whose length is the number of sentence clusters.
A non-zero entry corresponds to an occurrence of that sentence cluster
in the patient's corpus during the specified time period.}

\begin{figure}[h]
\centering
\includegraphics[height=7.5cm]{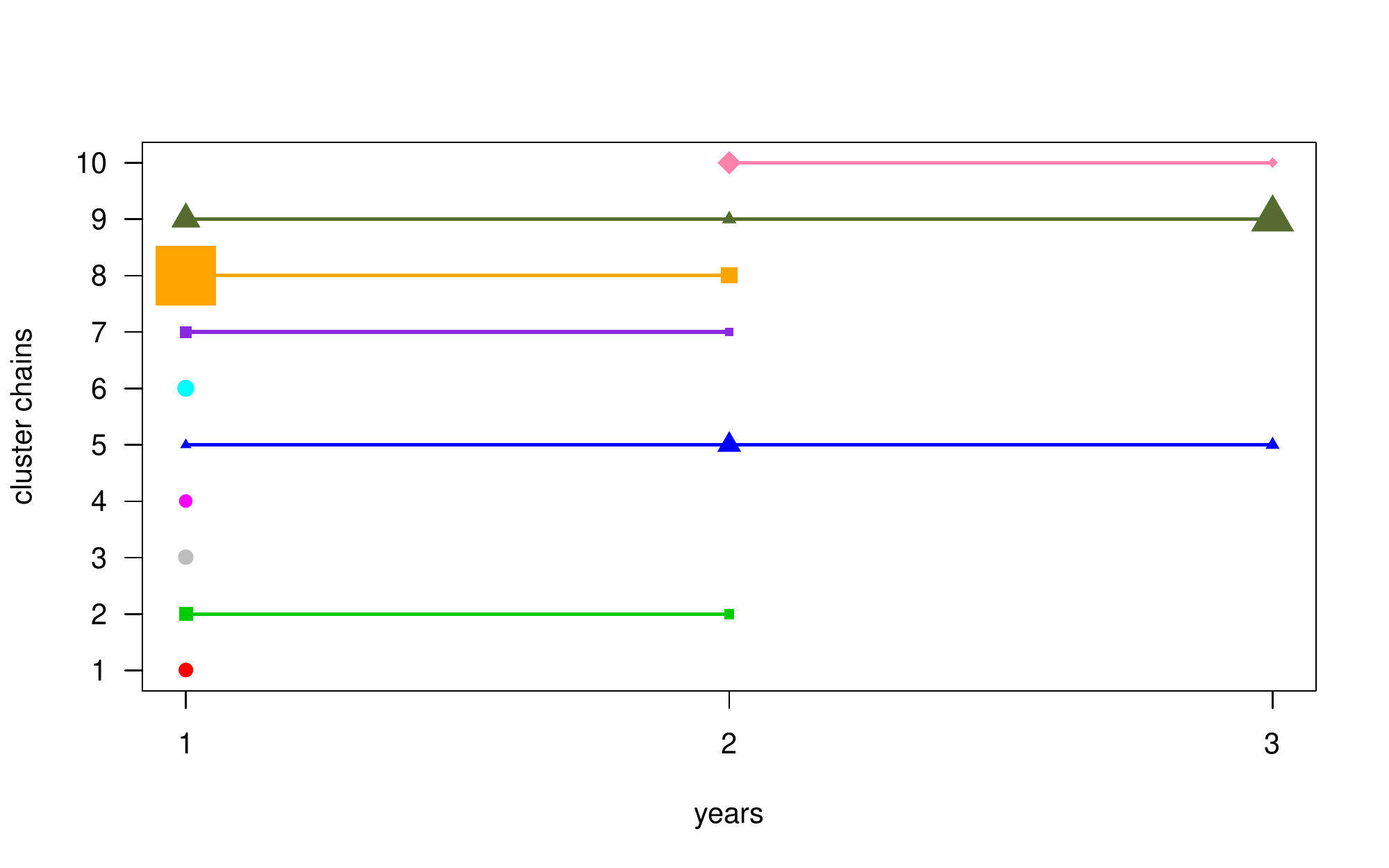}
\vspace{-5mm}
\caption{\label{EHR} Clusters over all 3 years of brain cancer
  patients. We find ten different cluster chains where 2 remain over
  al three years, 3 vanish after the second year and one new cluster
  comes up in year 2 and remains in the third year. Size of the tokens
  denote the cluster size, i.e.\ the number of patients per cluster.
  Note that patients can change clusters, so a cluster decreasing in size
  or disappearing does not necessarily mean those patients die or leave
  the hospital.}
\end{figure}

\begin{figure}[t]
  \begin{center}
    \vspace*{-3ex}
    \includegraphics[width=0.5\textwidth]{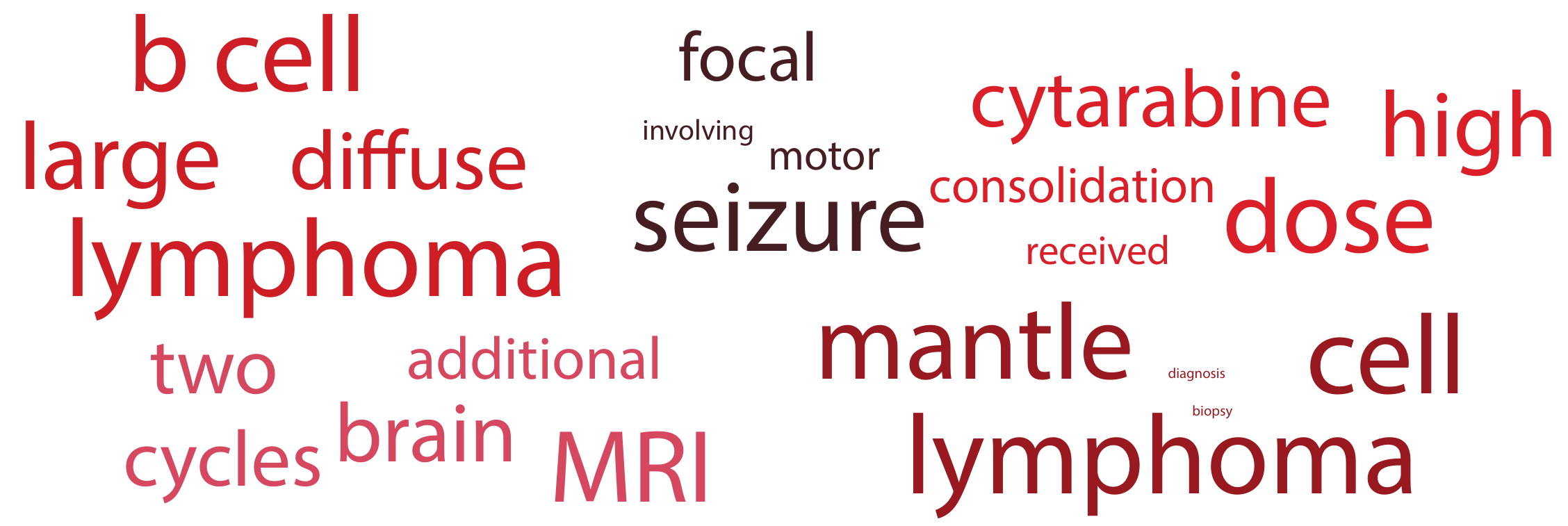}
    \vspace*{-3ex}
    \end{center}
  \caption{\label{cc1wc} Word clouds representing five sentence
    clusters that are observed in patients from cluster chain 1, the
    most deadly cluster. They describe patients that have blood
    cancers (lymphomas) in addition to brain cancer.\vspace*{-2ex}}
\end{figure}

\begin{figure}[h]
  \begin{center}
    \vspace*{-3ex}
    \includegraphics[width=0.5\textwidth]{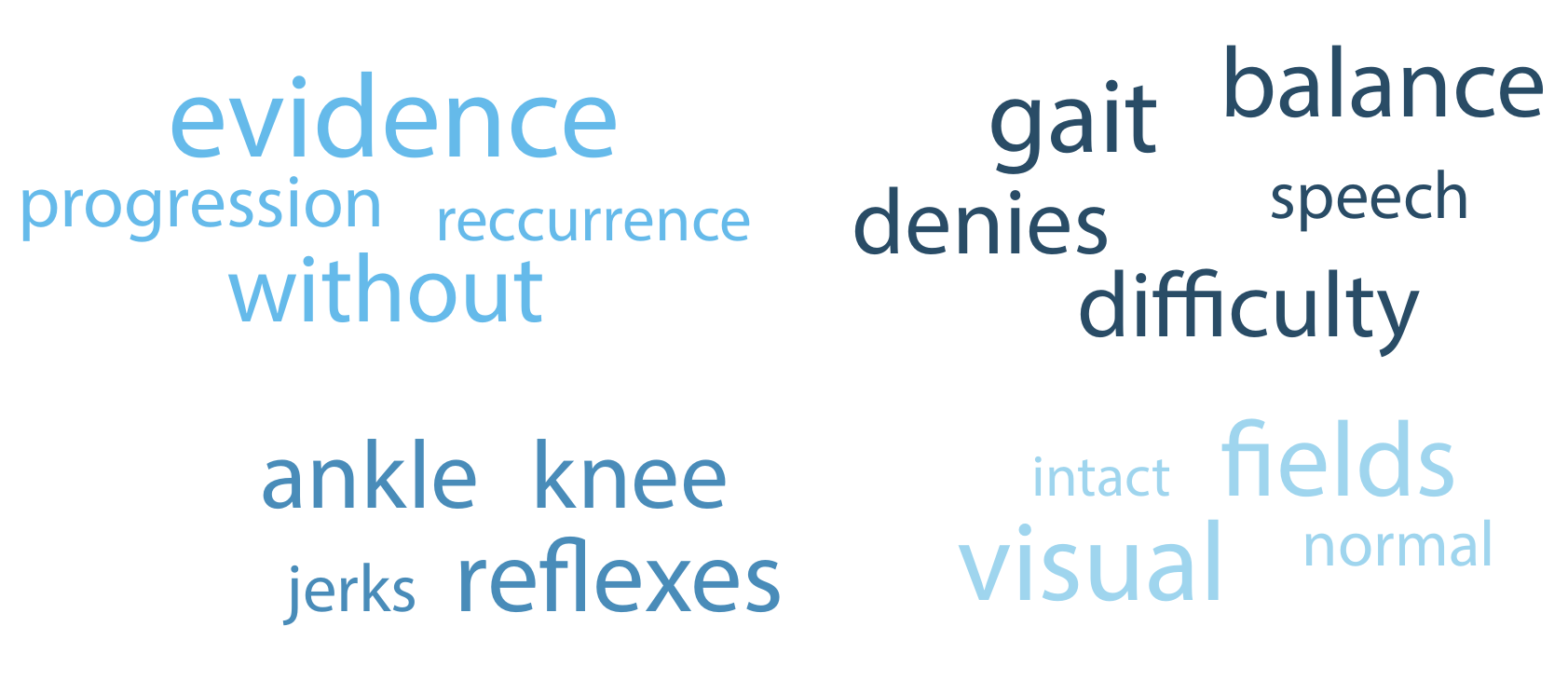}
    \vspace*{-3ex}
  \end{center}
  \caption{\label{cc5wc} Word clouds representing four sentence
    clusters that are observed in patients from cluster chain 5, the
    most positive cluster. These sentence clusters are "follow-up"
    language, such as checking reflexes or the ability to walk and see
    well. This indicates that the patients are in a relatively stable
    state under regular observation.\vspace*{-3ex}}
\end{figure}

\changedtext{In the first year, we have 704 patients, in the second
  year 170 and in the third year 123 patients. This data set has
  specific features which make our model particularly suitable to cope
  with this kind of data. First, the number of patients differs in
  every year. Second, patients disappear over the time course, either
  due to death or due to leaving the hospital. Third, patients do not
  necessarily need to have a document every year, so a patient can be
  absent from year 2 and appear in year 3. This gap occurred a total of
  31 times in our data set.
  This is why our flexible model is very well suited for this problem,
  as the model can deal with changing numbers of objects and changing
  number of clusters in every year, clusters can disappear or
  reappear, as well as patients. The result of our clustering model is
  shown in Fig.~\ref{EHR}.  On a standard computer, this
  experiment took roughly 6 hours, and the sampler stabilizes after
  roughly 500 sweeps.\\[0.5ex] 
  We observe ten different cluster chains over the time
  series. Note that patients can switch cluster chains over the
  years, as the tumor progresses, the status of the patient may
  change, resulting in more similarities to a different cluster chain
  than the year before.  To analyze the results of the method, we
  will discuss the most and least deadly clusters in more detail,
  as analyzing all subtleties between clusters would be out of the
  scope of this paper. \\[0.5ex]
  Cluster chain 1 is the most deadly cluster, with a death rate of
  80\%. Additionally, it only appears in the first year. Word clouds
  representing the sentence clusters of this patient group are shows
  in Fig.~\ref{cc1wc}. We can see that these patients are having seizures which
  indicates that the brain cancer is especially malicious.  They also
  show sentence clusters about two types of blood cancers, \emph{b
    cell} and \emph{mantle cell} lymphoma, and prescription of
  \emph{cytarabine}, which treats these cancers.  This combination of
  blood and brain cancers could explain why this cluster chain is so
  deadly.  } \\[0.5ex]
\changedtext{
  Cluster chain 5 is the least deadly chain with a death rate
  of 42\%. Word clouds representing sentence clusters for this patient
  group are show in Fig.~\ref{cc5wc}. }
\changedtext{These clusters consist of mainly "follow-up" language,
  such as checking the patients' gait, speech, reflexes and vision.
  The sentence clusters appear to indicate positive results,
  e.g.\ "Normal visual fields are intact", and "Patient denies
  difficulty with speech, language, balance or gait" are two prototype
  sentences representing two sentence clusters that appear in this
  chain.  Furthermore, there is a sentence cluster with prototype
  sentence "no evidence for progression," indicating that these
  patient's cancers are in a manageable state. }

\changedtext{Modeling patients over time provides important
  insights for automated analyses and medical doctors, as it is
  possible to check for every patient how the state of the patient as
  represented by the cluster membership changes over time. Also, if a
  new patient enters the study, one can infer, based on similarity to
  other patients, how to classify and possibly treat this patient best
  or to suggest clinical trials for each patient. Such clustering
  methods therefore make an important step towards solving the
  technical challenges of personalized cancer treatment.}

\section{Conclusion}
\label{conclusion}

In this work, we propose a novel dynamic Bayesian clustering model to
cluster time-evolving distance data.  A probabilistic model that is
able to handle non-vectorial data in form of pairwise distances has
the advantage that there is no need to embed the data into a vector
space. To summarize, our contributions in this work are five-fold: i)
We develop a dynamic probabilistic clustering approach that
circumvents the potentially problematic data embedding step by
directly operating on pairwise time-evolving distance data. ii) Our
model enables to track the clusters over time, giving information
about clusters that die out or emerge over time. iii) By using a
Dirichlet process prior, there is no need to fix the number of
clusters in advance.  iv) \changedtext{We test and validate our model
  on simulated data. We compare the performance of our new method with
  baseline probabilistic and hierarchical clustering methods. v) We
  use our model to cluster brain cancer patients into similar
  subgroups over a time course of three years.  Dynamic partitioning
  of patients would play an important role in cancer treatment, as it
  enables inference from groups of similar patients to an individual.
  Such an inference can help medical doctors to adapt or optimize existing
   treatments, assign billing codes, or predict survival times for a patient 
   based on similar patients in the same group.}

\begin{small}
  \paragraph{Acknowledgments}
  We thank Natalie Davidson, Theofanis Karaletsos and David Kuo for
  helpful discussions and suggestions.  JV and MK were partly funded
  through postdoctoral fellowships awarded by the Swiss National Science
  Foundation (SNSF; under PBBSP2\_146758) and by the German Research
  Foundation (DFG; under Kl 2698/1-1 and VO 2003/1-1), respectively. We
  gratefully acknowledge funding from Memorial Sloan Kettering Cancer
  Center and the National Cancer Institute (grant
  1R01CA176785-01A1). Access to patient data is covered under IRB Waiver
  \#WA0426-13.
\end{small}

\bibliographystyle{plain}

%\GR{can you please find a place where to cite our first text analysis
%  paper \cite{Chan13}.} 

\bibliography{all_ref}

\end{document}